\documentclass[11pt]{article}

\usepackage[T1]{fontenc}
\usepackage[a4paper,margin=2.5cm]{geometry}
\usepackage{graphicx}
\usepackage{multirow}
\usepackage{amsmath,amssymb,amsfonts}
\usepackage{amsthm}
\usepackage{mathrsfs}
\usepackage[title]{appendix}
\usepackage{xcolor}
\usepackage{textcomp}
\usepackage{booktabs}
\usepackage{listings}
\usepackage{float}
\usepackage[super]{nth}
\usepackage{tabularx}
\usepackage{caption}
\usepackage{subcaption}
\usepackage{acronym}
\usepackage{tikz}
\usepackage{pgfplots}
\pgfplotsset{compat=1.18}

\usepackage[linesnumbered,ruled,vlined]{algorithm2e}
\usepackage{algpseudocode}
\usepackage{algorithmicx}
\usepackage{algcompatible}
\usepackage[normalem]{ulem}

\usepackage[numbers,sort&compress]{natbib}     % [1,2] 形式的数字引用，正常字号
\usepackage{authblk}
\usepackage{url}
\usepackage[colorlinks=true,citecolor=blue,linkcolor=blue,urlcolor=blue]{hyperref}   % 引用、交叉引用、URL 均为蓝色文字，无边框
\usepackage{cleveref}                          % 必须在 hyperref 之后加载

\usepackage{fancyhdr}

\theoremstyle{plain}

\theoremstyle{remark}

\theoremstyle{definition}

\providecommand{\burl}[1]{\url{#1}}

\begin{document}

\title{Deep learning enables urban change profiling through alignment of historical maps}

\author[1]{Sidi Wu}
\author[1]{Yizi Chen$^{*}$}
\author[2]{Maurizio Gribaudi}
\author[3]{Konrad Schindler}
\author[4]{Clément Mallet}
\author[4]{Julien Perret$^{\dagger}$}
\author[1]{Lorenz Hurni$^{\dagger}$}

\affil[1]{Institute of Cartography and Geoinformation, ETH Zurich, Switzerland}
\affil[2]{CRH, École des hautes études en sciences sociales (EHESS), France}
\affil[3]{Photogrammetry and Remote Sensing, ETH Zurich, Switzerland}
\affil[4]{LASTIG, Univ Gustave Eiffel, IGN, Géodata Paris, France}

\date{\small
  $^{*}$Corresponding author: \href{mailto:yizi.chen@ethz.ch}{yizi.chen@ethz.ch}\\
  $^{\dagger}$These authors jointly supervised this work.}

\maketitle
\thispagestyle{fancy}

\begin{abstract}
Prior to modern Earth observation technologies, historical maps provide a unique record of long-term urban transformation and offer a lens on the evolving identity of cities.
However, extracting consistent and fine-grained change information from historical map series remains challenging due to spatial misalignment, cartographic variation, and degrading document quality, limiting most analyses to small-scale or qualitative approaches.
We propose a deep-learning-based framework for fine-grained urban change analysis from large collections of historical maps, built on a modular design that integrates spatial map alignment, multi-temporal object extraction, and change estimation.
This framework shifts the analysis of historical maps from ad hoc visual comparison toward systematic, quantitative characterization of urban change. 
Experiments demonstrate the robust performance of the proposed alignment and object extraction methods.
Applied to Paris between 1868 and 1937, the framework reveals the spatial and temporal heterogeneity in urban transformation, suggesting its potential relevance for research in the social sciences and humanities. 
The modular design of our framework provides flexibility for future adaptation to different cartographic contexts and downstream applications.
\end{abstract}

\renewcommand{\figurename}{Figure}
\section{Introduction}\label{sec:intro}
The last two centuries have witnessed unprecedented modernization and urbanization. 
Analyzing urban change is pivotal to understanding these long-term dynamics, as cities continuously evolve in response to infrastructural expansion, shifting demographics, and policy decisions.
By examining how cities transform across space and times, researchers can uncover mechanisms that shape urban development, thereby developing more sustainable and resilient development strategies~\citep{ogle2017quantifying, larkham2019persistence, tumturk2024measuring, chakraborty2022pursuit, wu2023urbanization}.
Most studies of urban change focus on the past 50 years~\citep{liu2020change,yang202130,huang202130}, largely because of the availability of satellite imagery and digital geographic records. 
Yet many of the most consequential transformations occurred earlier, especially during the 19\textsuperscript{th} and early 20\textsuperscript{th} centuries, when modern planning principles were first introduced.
For instance, the transformation of Paris during the second French Empire (so-called Haussmannian transformations) in the mid-19\textsuperscript{th} century was a pioneering and influential project, with effects that extended well into the 20\textsuperscript{th} century. This project reshaped Paris from a congested medieval city into a modern metropolis by introducing wide boulevards, harmonized building façades, improved sanitation, and expanded public spaces such as parks and squares.
Another pioneer project, the Eixample plan of Barcelona, initiated in the mid-19\textsuperscript{th} century, introduced a grid-based layout with wide, uniform streets, abundant open spaces, and mixed-use building blocks that balanced practical functionality with aesthetic appeal.
A further example is the unification of Switzerland as a federal state in 1848, which accelerated nationwide railway construction, spurred urban growth, and supported the development of public infrastructure.
Characterizing patterns of change in the early modernization period provides us a crucial historical lens and a more holistic view of the cities' identity as a continuation of past transformations, rather than a series of isolated developments.

Thanks to advances and large-scale applications of geodetic surveying and cartography methods in the last two centuries, topographic and cadastral maps have been produced with high geometric accuracy, rich detail, and regular updates every decade in response to ongoing landscape and urban changes. 
For example, Atlas Municipal, a collection of city planning maps of Paris and its suburbs, meticulously documented buildings, building blocks, and other urban infrastructures such as parks and public spaces in high resolution from the mid-19\textsuperscript{th} century to the early 20\textsuperscript{th} century, providing a valuable record of the early modernization efforts.
Siegfried Maps, the topographic atlas of Switzerland, produced between the mid-19\textsuperscript{th} and mid-20\textsuperscript{th} centuries, offer a comprehensive and precise representation of Swiss territory, shaped by national unification and infrastructural growth.
Analog maps, as the only spatially explicit source available before the advent of modern Earth observation techniques, have been scanned, geo-referenced, and digitally archived in recent years. Notable collections include the USGS Topographic Maps~\cite{usgs_topoview} and the David Rumsey Map Collection~\cite{david_rumsey}. These historical maps have been increasingly utilized to reconstruct past landscapes, analyze long-term spatial transformations, and study urban and environmental evolution~\citep{sanantonio2014urban, Burghardt2022road, Tonolla2021decades, PICUNO2019investigating}.

\begin{figure}[htbp]
    \centering
    \includegraphics[width=\textwidth]{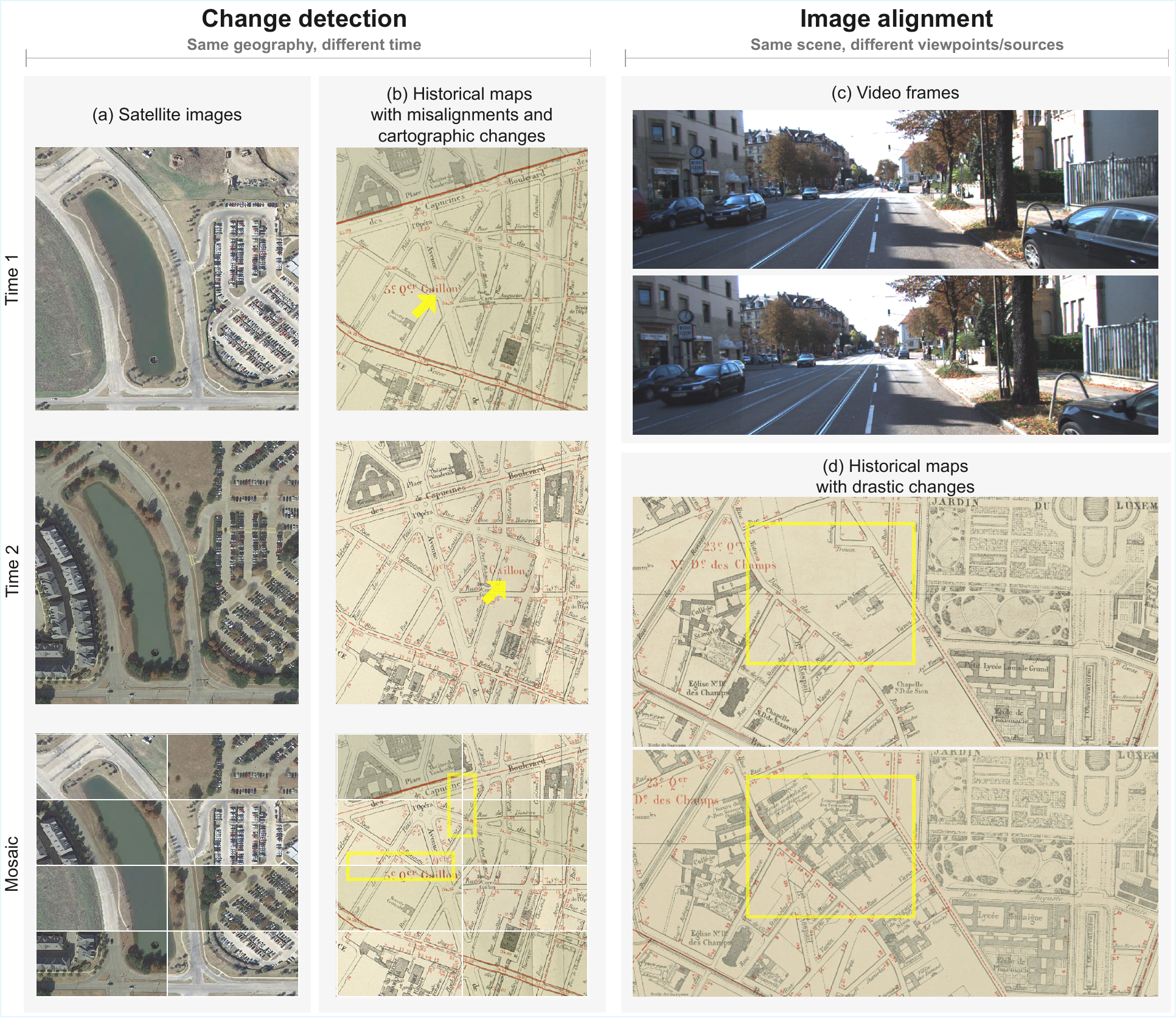}
    \caption{Historical maps violate assumptions underlying standard change-detection and image-alignment methods. (a) Two satellite images~\citep{Chen2020levircd} with a time difference of 5-10 years exhibit stable object locations across years, allowing simple overlays to reveal changes. (b) Historical maps~\cite{villeparis1878atlas, villeparis1888atlas} from editions a decade apart display obvious object misalignment: object locations shift across sheets (yellow boxes at mosaic boundaries), making direct comparison unreliable. In addition, cartographers frequently adjust text labels for layout or legibility (yellow arrows), introducing displacements unrelated to real-world change. (c) Video frames~\citep{geiger2013kitti} contain smooth, coherent motions where dense-matching algorithms are typically used to align frames, track objects, and estimate per-pixel motion trajectories. (d) Conversely, historical maps exhibit substantial content changes (yellow boxes)~\cite{villeparis1888atlas, villeparis1900atlas}. These drastic changes, together with deliberate cartographic layout adjustments, undermine the assumptions of dense-matching algorithms, i.e., stable object correspondences for accurate image alignment.}
    \label{fig:change_alignment_comparison}
\end{figure}

Traditionally, detecting changes from historical maps has relied on manual, qualitative, or semi-automatic inspection~\citep{sanantonio2014urban, PICUNO2019investigating,Tonolla2021decades}, or has focused on aggregate statistics to characterize broader spatial patterns.~\citep{Burghardt2022road, rath2023settlement}. In recent years, deep-learning-based approaches have been explored to automate feature extraction for buildings, roads,  rivers, and forests~\citep{wu2022leveraging, chen2024benchmark, wu2023domain, Heitzler2020building, Uhl2020building}. However, extracted features often misalign across time, even after being geo-referenced/-located in the same coordinate system, primarily due to surveying errors, cartographic generalization, variations in production methods, storage-induced distortions, and changes in feature signatures over time~\citep{wu2022unsupervised}. This fails to capture fine-grained changes at specific past dates and prevents reliable qualitative and quantitative analyses over long time frames.
Despite advancements in neural-network-based change detection~\citep{daudt2018fully, peng2019end, chen2021Transformer, Fang2022snunet, ZHANG2020bitemp, hou2024war}, most existing methods are designed for multi-temporal satellite images, where objects are typically assumed to occupy the same position across time---an assumption that does not hold for historical maps, illustrated in~\Cref{fig:change_alignment_comparison} (a)(b). Moreover, these approaches usually require predefined concepts of ``change'', which can be subjective and strongly dependent on research questions, spatial scale, and temporal context. 
For historical maps, changes are often subtle and the ``truth'' of past transformations is uncertain, imposing such rigid definitions can overlook important nuances. This highlights the need for more flexible approaches.

Detecting changes from historical maps therefore require precises spatial alignment of objects, robust object detection methods that can handle graphic variations across decades, and a flexible way of categorizing change.
While many studies have developed neural networks to estimate how objects shift between images at the pixel level~\citep{shen2020ransac,melekhov2019dgc,truong2020glu,truong2023pdc,truong2021warp, park2022simsac,baker2011database, hur2020optical, dosovitskiy2015flownet, sun2018pwc, sun2019models, hui2018liteflownet}, often referred to as dense matching, 
aligning historical map objects presents unique challenges (\Cref{fig:change_alignment_comparison}): the underlying city can change substantially, and cartographers frequently shift text labels and symbols for layout or legibility reasons, complicating correspondence estimation.
We therefore argue that a suitably designed, yet not over-engineered, deep learning approach is necessary to leverage information available and retrieve objects in each map, while ensuring consistency and spatial alignment through the full series. We assume that agnostic change profiling can be decomposed into two independent components: spatial alignment and fine-grained object extraction. Further quantitative or qualitative analyses addressing specific historical or urban research questions require dedicated spatial analysis tools and domain expertise and are therefore beyond the scope of this work.
To address the mentioned challenges, we embed explicit cartographic guidance into the alignment network during training in a resource-efficient way. To make object detection robust to graphic variations, particularly where older maps suffer from quality degradation, we incorporate multi-temporal information from neighbouring years, allowing the model to resolve ambiguities using additional context. Together, these rigorous alignment and object detection across time enable us to automatically quantify changes by simply comparing overlaps between corresponding objects.

We propose a fully automated modular framework for systematically tracking and measuring detailed changes in historical maps over space and time, with components designed to address technical challenges associated with map alignment and object extraction. The framework reduces the need for exhaustive manual delineation and ad hoc visual comparison, and produces map objects that can be compared consistently across years. We demonstrate its effectiveness by tracking urban changes associated with early modernization in Paris between 1868 and 1937 as an end-to-end example.
The resulting aligned objects and change maps provide a structured basis for detailed studies of urban change, enabling comparative, long-term analyses that can support a broad range of historical, urban, and environmental research.
In addition, we demonstrate the applicability of the modular framework to other map series and related applications, such as distortion analysis and different levels of change analysis, ranging from fine-grained structural changes to broader morphological changes.

\section{Results}
\subsection{Overview of the framework}

\begin{figure}[htbp]
    \centering
    \includegraphics[width=\textwidth]{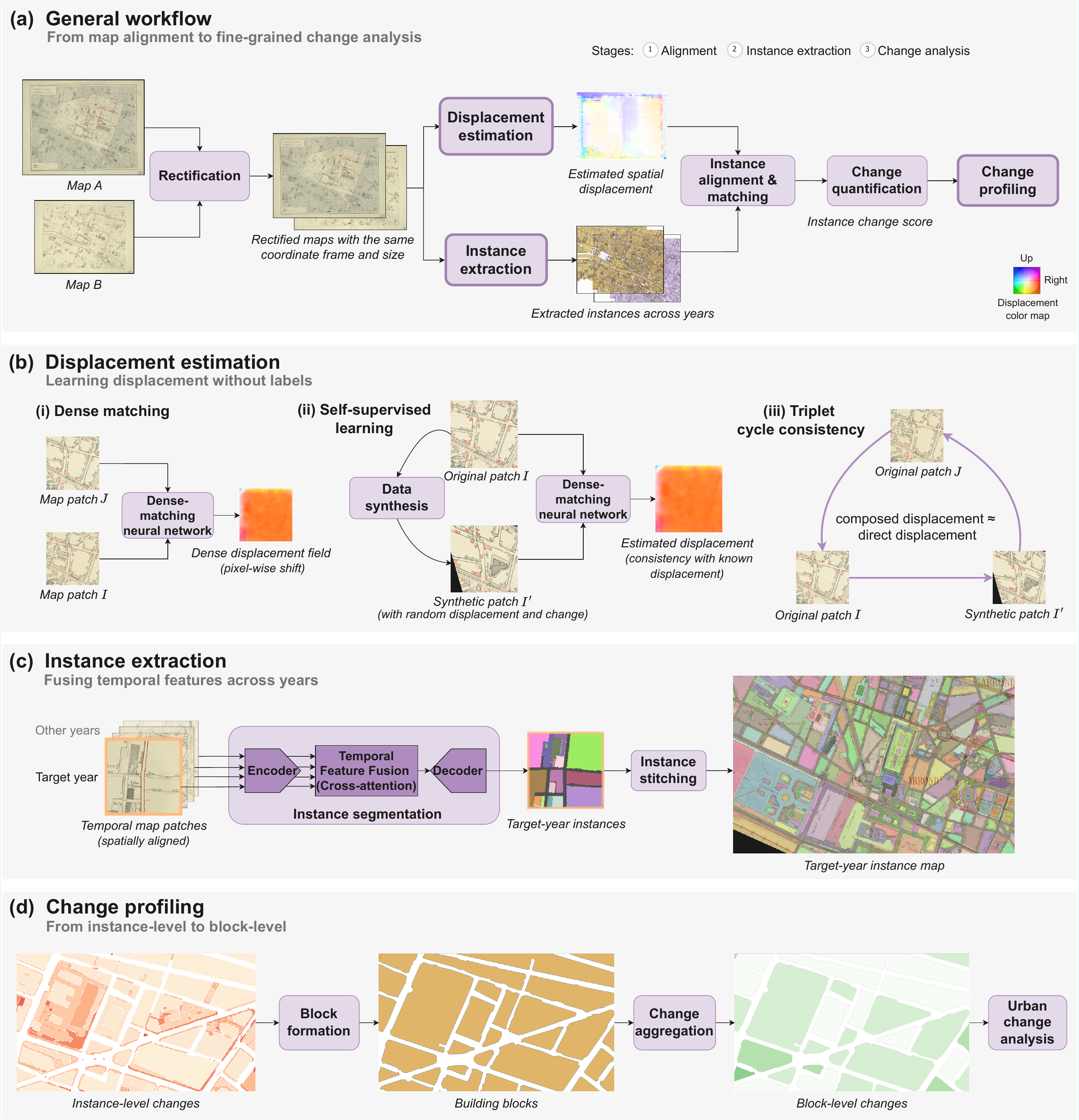}
    \caption{Method overview.
    (a) Workflow for historical map alignment and change analysis. Maps covering the same location are first rectified to a common image plane. For map pairs from different editions (years), we estimate a dense displacement field for pixel-level alignment and extract urban instances for each edition. The displacement field is used to align and match instances across years, enabling fine-grained instance-level change analysis. 
    (b) Dense displacement estimation without ground-truth correspondences. A dense-matching network is trained using self-supervision from synthetic transformations with known displacements, and refined with a triplet cycle-consistency constraint linking real and synthetic views. 
    (c) Instance extraction with temporal knowledge. A shared encoder--decoder integrates information from multiple years via temporal feature fusion to predict instance segmentation for the target year. The patch-wise predictions are stitched to produce a complete instance map. 
    (d) Instance-level changes are aggregated into urban blocks, a fundamental spatial unit in urban morphology and planning, to derive block-level change profiles. These profiles enable planning-relevant interpretation of long-term urban morphological change.}
    \label{fig:overview}
\end{figure}

Historical maps provide unique records of long-term urban evolution, but their analysis is fundamentally challenged by severe geometric misalignments and significant diversity (e.g., cartographic variations). To address these challenges, we introduce a unified computational framework that enables dense alignment and fine-grained change detection across historical map editions.
An overview of the proposed pipeline is shown in \Cref{fig:overview}.
Our framework consists of three main stages: map alignment, instance-level extraction, and fine-grained change analysis. 
The first two steps are designed to be agnostic to any quantitative and qualitative analysis, while the last one is presented here to illustrate the high relevance of such generic outputs for social and human sciences.
As shown in \Cref{fig:overview}~(a), given geo-referenced historical map archives, we first rectify maps depicting the same locations into a common image plane by unifying spatial resolution, cropping to the same extent, and converting geographical coordinates into pixel coordinates with a shared origin. This enables subsequent dense alignment across map editions and suitable qualitative and quantitative assessment for thematic end-users.
We then estimate dense displacement fields between map pairs and use them for precise pixel-wise alignment. Based on the aligned maps, an instance detection network identifies urban infrastructures such as buildings, building blocks, and parks. Changes are detected by directly comparing instance geometries across time.

\Cref{fig:overview}~(b) illustrates the displacement estimation component of our framework. Since ground-truth displacement fields between historical maps are not available, we adopt a self-supervised deep learning training strategy inspired by~\cite{truong2021warp}. During training, we synthesize paired images with known displacement fields by applying combinations of geometric transformations, which are used to supervise the displacement estimation deep-learning-based network. 
To adapt the model to real historical maps, we further refine displacement predictions by enforcing triplet cycle consistency across multiple images, including both the real image pair and the synthetically warped image.  The real image pair consists of small patches cropped from maps of different years at the same geographic location.
To explicitly account for the challenges posed by historical maps, we simulate drastic object changes and cartographic modifications of text labels during training. The~\Cref{sec:disp_results} shows the effectiveness of the proposed displacement estimation and map alignment method.

For instance extraction, we extract building units as individual instances for small map patches (\Cref{fig:overview}~(c)).
To improve robustness against graphic defects and local ambiguities, we incorporate temporal information from multiple map editions. After aligning maps using the estimated displacement field, we fuse features from corresponding locations across time using a temporal integration mechanism inspired by~\cite{wu2023cross}. Patch-wise predictions are subsequently merged to produce complete instance maps for each year.
\Cref{sec:instance_extraction} presents the effectiveness of our instance extraction method.

While instance-level comparisons enable detailed change detection, individual instances may represent either entire buildings or partial building elements, which can fragment change analysis.
To make the results more interpretable and coherent, we aggregate detected changes at the level of building blocks, which are more meaningful spatial units aligned with the interpretive practices of urban historians and urban planners (\Cref{fig:overview}~(d)). To this end, we derive the change profiles of Paris from 1868 to 1937, as demonstrated in \Cref{sec:change_profiles}. While our main studies were conducted on the Atlas Municipal map series for Paris, we also present additional results on maps from other countries.

\subsection{Accuracy of dense map alignment}
\label{sec:disp_results}

Our self-supervised displacement estimation method extends the WarpC-GLU-Net framework~\citep{truong2021warp} by introducing a self-synthesis strategy that integrates cartographic knowledge, enabling learning from large archival collections without ground-truth displacement annotations.
We evaluate alignment performance both within the training area and in previously unseen regions to assess spatial generalizability.
We compare our method with two state-of-the-art dense image-matching approaches, GLU-Net~\citep{truong2020glu} and WarpC-GLU-Net (hereafter WarpC), designed to handle large displacements through self-supervised learning strategies. WarpC serves as a strong baseline, as it does not incorporate cartographic simulations, allowing us to isolate the impact of our domain-specific adaptations. More details of the state-of-the-art dense image-matching approaches can be found in Supplementary Note 4.

In the absence of per-pixel ground-truth, we evaluate performance indirectly by assessing alignment at the image and object levels, after applying the estimated displacement fields.
At the image level, we measure the similarity between aligned maps using the Structural Similarity Index Measure (SSIM). At the object level, we evaluate alignment accuracy using the Chamfer Distance (CD) between corresponding objects. To reduce the influence of genuine urban changes and outliers, we exclude the top 10\%, 20\%, and 40\% of CD values: the large object distances can be treated as indicators of true change.
Beyond alignment accuracy, we assess the smoothness and temporal consistency of the estimated displacement fields, which are critical for stable change analysis. Spatial smoothness is quantified using the mean variation (mV), defined as the average absolute gradient between neighboring pixels. Temporal consistency is evaluated by computing the L1 difference between displacement fields estimated from consecutive year triplets (e.g., 1900–1912–1925).

\begin{figure}
    \centering
    \includegraphics[width=\textwidth]{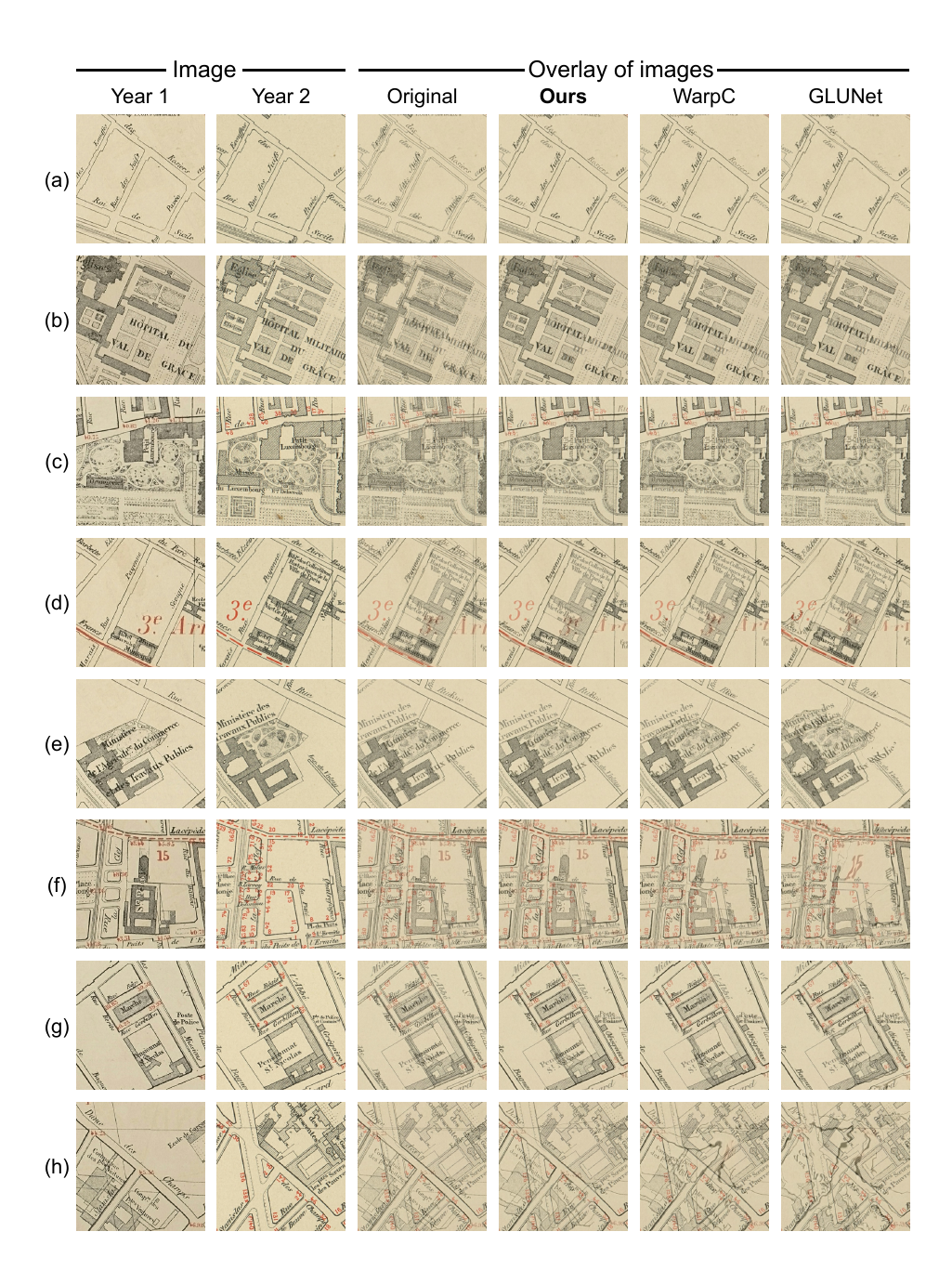}
    \caption{A qualitative comparison of image alignment across different methods. In (a-c), all methods effectively align images despite large displacements between the originals. Our method shows robustness to text displacements in (d) and (e), as well as to object changes in (f-h), maintaining stability even in extreme cases, as seen in (e). }
    \label{fig:comparison_qualitative}
\end{figure}

As summarized in \Cref{tab:flow-estimation}, all methods substantially reduce object misalignment in both training and unseen test areas, demonstrating strong performance and spatial generalizability. 
Compared to WarpC, our method achieves comparable alignment accuracy, while producing markedly smoother displacement fields with substantially higher temporal consistency.
These properties are essential for minimizing jitter and ensuring reliable downstream change quantification.
Qualitative results further support these findings (\Cref{fig:comparison_qualitative}). While all methods successfully align maps with large initial displacements (a-c), competing approaches exhibit noticeable distortions when text elements are randomly displaced (d–e), or when structural changes occur (f–g). In contrast, our method remains stable under both conditions, maintaining coherent alignments even in the presence of extreme cartographic and structural variations (h). Qualitative comparison of estimated displacement fields can be found in Supplementary Figure 4. 
Additional tests on other map datasets can be found in Supplementary Table 1 and Supplementary Figure 5. Our model still achieves the best performance, especially in improving the smoothness and temporal consistency on these data.

\begin{table}[tb]
\centering
\caption{Model performance for displacement estimation in both training areas (15 districts) and unseen testing areas (5 districts). 
We measure the CD by removing the 10\%, 20\%, and 40\% largest distances. 
Smoothness is evaluated by mV, and temporal consistency by L1 within triplets from consecutive years. 
Arrows indicate the direction of better performance. The best values are marked in bold.}
\label{tab:flow-estimation}

\setlength{\tabcolsep}{6pt}
\renewcommand{\arraystretch}{1.2}

\begin{tabular}{l c c c c c c}
\toprule
 & Image & \multicolumn{3}{c}{Object} & Disp. & Temp. Cons. \\
\cmidrule(lr){2-2} \cmidrule(lr){3-5} \cmidrule(lr){6-6} \cmidrule(lr){7-7}
Method
& SSIM$\uparrow$ 
& CD(10\%)$\downarrow$ 
& CD(20\%)$\downarrow$ 
& CD(40\%)$\downarrow$ 
& mV$\downarrow$ 
& L1$\downarrow$ \\
\midrule

\multicolumn{7}{c}{\textsc{Training Area}} \\
\midrule
Unaligned   & 0.72 & 5.50 & 3.96 & 1.89 & --   & --   \\
GLU-Net     & 0.78 & 2.57 & 1.73 & 0.82 & 0.66 & 7.16 \\
WarpC       & 0.83 & 1.47 & 0.95 & 0.44 & 0.29 & 3.87 \\
\textbf{Ours}
            & 0.83 & \textbf{1.44} & \textbf{0.93} & \textbf{0.43} & \textbf{0.23} & \textbf{2.83} \\
\midrule

\multicolumn{7}{c}{\textsc{Testing Area}} \\
\midrule
Unaligned   & 0.80 & 3.29 & 1.54 & 0.31 & --   & --   \\
GLU-Net     & 0.85 & 2.12 & 1.45 & 0.68 & 0.54 & 5.17 \\
WarpC       & 0.88 & 1.20 & 0.80 & 0.35 & 0.23 & 2.46 \\
\textbf{Ours}
            & 0.88 & \textbf{1.19} & \textbf{0.79} & \textbf{0.34} & \textbf{0.19} & \textbf{1.73} \\
\bottomrule
\end{tabular}
\end{table}

Beyond its role in map correspondence, the alignment output was also examined for downstream uses. Supplementary Figure 10 illustrates how the estimated displacement fields can characterize geometric differences between map editions for map distortion analysis. Alignment-derived correspondences were additionally used for label propagation (Supplementary Figure 11), reducing manual annotation time in the Paris case from approximately 70 hours to 8 hours per sheet.

\subsection{Instance extraction performance}
\label{sec:instance_extraction}
Reliable instance extraction is essential for downstream change analysis. We adopt Mask2Former~\citep{cheng2021mask2former}, a state-of-the-art transformer-based instance segmentation framework, as our baseline to identify building units---defined as closed-shape polygons in our approach. Mask2Former already performs well on historical maps. Still, graphic defects and missing details can degrade performance.
To improve robustness, we integrate multi-temporal information from multiple map editions, leveraging complementary cues across time, mitigating local ambiguities, and improving the temporal consistency across editions.

\begin{figure}[tb]
    \centering
    \includegraphics[width=0.9\textwidth]{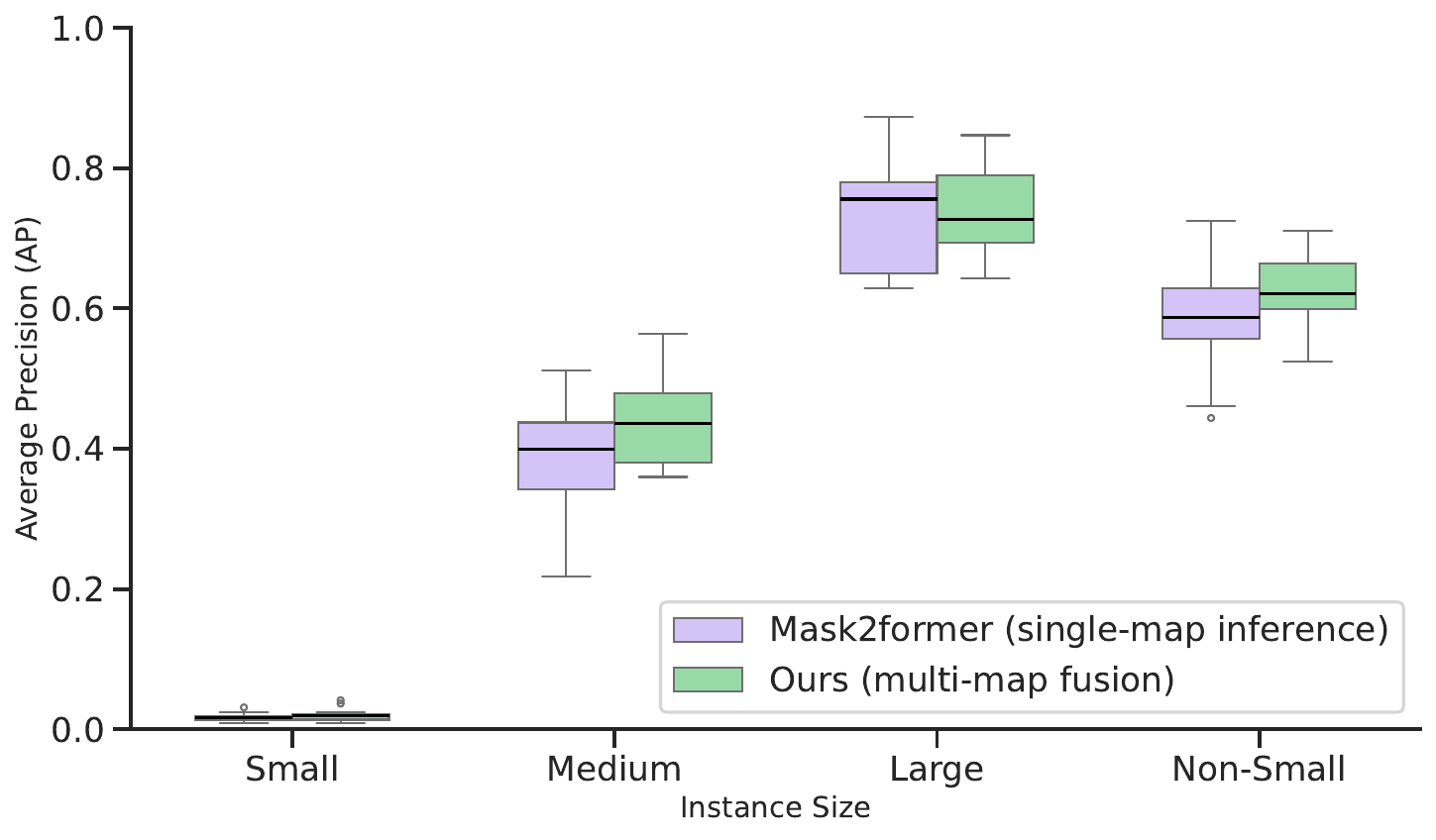}
    \caption{Average precision (AP) scores for objects categorized by area: small (area $\leq 500$ m$^2$), medium ($500$ m$^2<$ area $\leq 2000$ m$^2$), large (area $>2000$ m$^2$), and non-small (area $>500$ m$^2$). We compare single-map segmentation (Mask2Former as the baseline) with our proposed multi-map fusion strategy. For each box plot, $n=12$ test areas were used to derive the displayed statistics, with each test area contributing one AP value. The test areas span two years, and each contains 400–3000 object instances. Box plots show the median as the center line, with the lower and upper bounds of the boxes corresponding to the 25th and 75th percentiles, respectively. Whiskers extend to the smallest and largest observations within 1.5 times the interquartile range (IQR) from the lower and upper quartiles. Observations beyond the whiskers are shown individually as outliers. Our model improves overall accuracy and reduces the IQR and number of outliers.}
    \label{fig:instance_seg}
\end{figure}

We evaluate instance extraction performance using average precision (AP), computed by stacking predictions across all available years to jointly assess spatial accuracy and temporal stability. \Cref{fig:instance_seg} compares AP scores obtained using single-map inference with those from our multi-temporal fusion strategy, stratified by object size. While Mask2Former struggles to detect small objects (area < 500~m$^2$), it performs reliably on medium, large, and non-small categories, far more relevant for urban change analysis. Accordingly, we focus on objects larger than 500~m$^2$ in subsequent experiments.
Incorporating multi-temporal information consistently improves AP and reduces variability across years, as reflected by a smaller interquartile range and fewer outliers. These improvements are particularly pronounced for medium-sized buildings, which are more susceptible to graphic degradation and partial occlusion.

\begin{figure}
    \centering
    \includegraphics[width=0.9\textwidth]{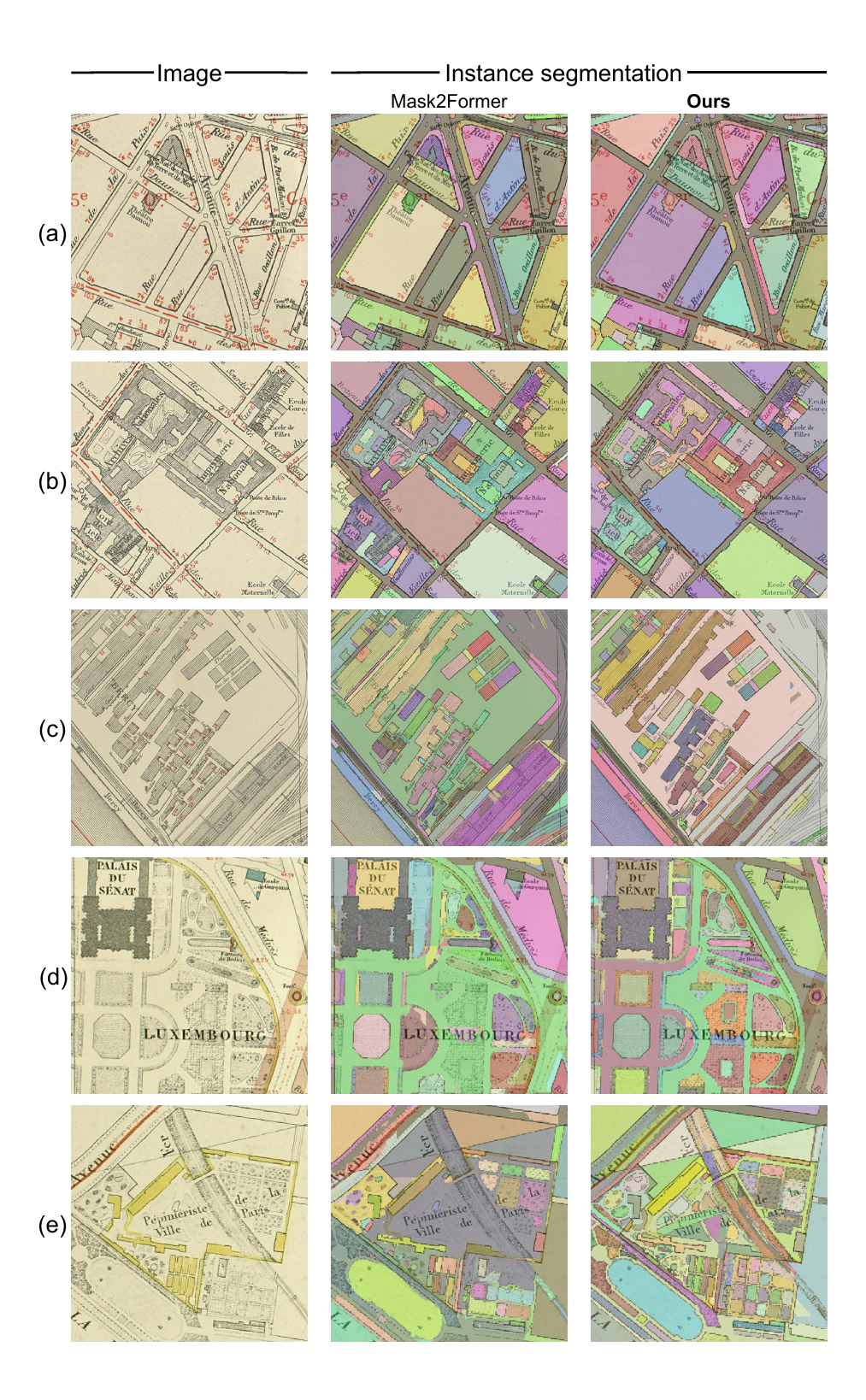}
    \vspace{-0.2cm}
    \caption{Qualitative results of our instance extraction method. Individual instances are shown in distinct colors. Our model performs comparably or slightly better to the baseline Mask2Former in (a) building blocks and sidewalks, (b) intricate building structures, and (c) railway neighborhoods, while demonstrating clearly improved performance in detecting faded features such as buildings and gardens (d,e).}
\label{fig:instance_segmentation}
\end{figure}

Qualitative examples are shown in \Cref{fig:instance_segmentation}. While the baseline Mask2Former already handles open-border structures such as sidewalks in (a) effectively, our approach demonstrates improved robustness particularly in challenging scenarios where building outlines are faded or partially missing in (d) and (e). This increased stability is critical for reducing spurious change detections arising from differences in map quality across years.
For subsequent change analysis, individual building units are aggregated into building blocks by uniting connected instances and applying morphological post-processing to remove small holes and filter out blocks below an area threshold of 500~m$^2$.  Additional tests on other maps can be found in Supplementary Table 2 and Supplementary Figure 6. Both the Mask2Former baseline and our fusion strategy perform similarly well on instance segmentation of Siegfried maps, achieving higher AP than Atlas Municipal due to the relatively lower complexity of building recognition.

\subsection{Fine-grained change profiles of Paris}
\label{sec:change_profiles}
We aggregated individual building instances into building blocks (or urban blocks) delimited by major infrastructure elements like roads, rivers, and former fortification boundaries, following the definition of~\cite{chazalon2021icdar}. This broader interpretation of blocks also allows infrastructure-dominated units, such as railway areas, to be included when they form coherent enclosed urban entities.
We quantified temporal change using an Intersection over Union (IoU)–based indicator.  
Such an aggregation is made possible since buildings are exactly delineated: the adequate scale change (blocks) and the detection of the road network are now possible and consistent through the epochs. 
For each pair of consecutive years, instances from aligned map sheets were matched using a one-to-one Hungarian assignment, and IoU values were computed for each matched pair. The block-level change indicator was then obtained by averaging IoU scores within each block. Lower values indicate greater structural transformation and thus serve as a proxy for urban change intensity.

\begin{figure}
    \centering
     \includegraphics[
        width=\linewidth,
        trim=0 0.8cm 0 0,
        clip
    ]{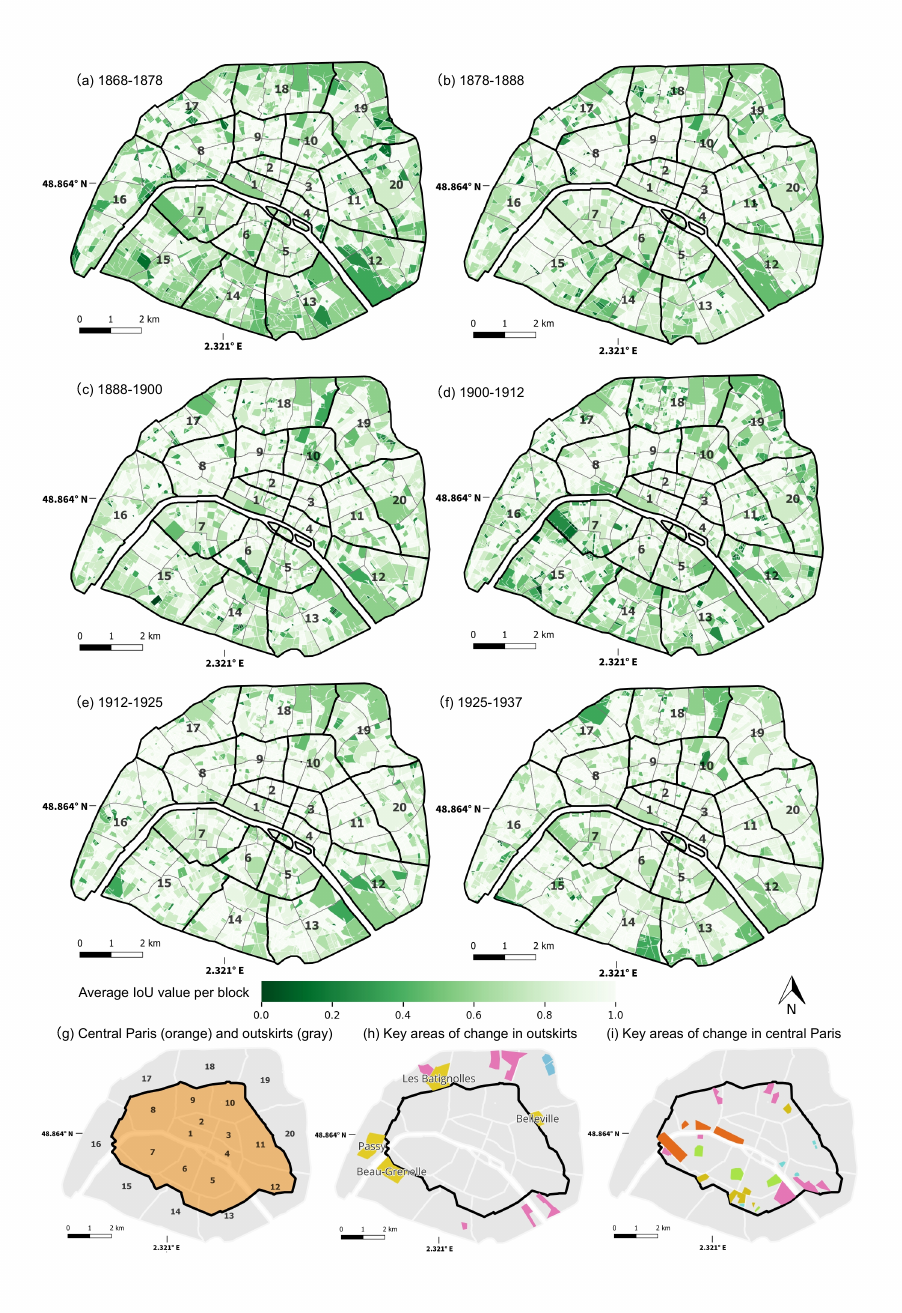}
    \caption{Temporal change patterns of Paris from 1868--1937. Each map covers 20 arrondissements (see (g)), subdivided into four quartiers each, for a total of 80 quartiers. Each map displays the average Intersection over Union (IoU) values per building block, calculated from the overlaps of all aligned instances within that block across two consecutive years, serving as a geometric proxy for changes---lower values indicate larger changes potentially. (h) highlights areas in the outskirts associated with rail freight (pink) and waterborne freight (blue), as well as areas that remained relatively stable (yellow). (i) highlights changed areas in central Paris, including hospitals (yellow), parks (green), railway stations (pink), world exposition sites (orange), and other public buildings (blue).}
    \label{fig:paris_changes}
\end{figure}

Across the full study period, change intensity is highly uneven in both space and time in~\Cref{fig:paris_changes}. The decade spanning 1868–1878 stands out as the period of most intensive transformation, with systematically lower IoU values than in subsequent periods. This phase corresponds to a time of major reorganization following the annexation of peripheral areas to Paris in 1860, shown in gray in
\Cref{fig:paris_changes}~(g).
These areas, also called Paris Zone (zone suburbaine or petite banlieue)~\citep{cannon2016paris}, correspond to the zone between the Wall of the Fermiers généraux (black line) and the Thiers enclosure (the extent of the map). 
They exhibit stronger and more persistent restructuring throughout the study period than central Paris (the city prior to 1860), shown in orange in 
\Cref{fig:paris_changes}~(g).
By contrast, central Paris underwent relatively limited modifications and remained mostly stable after 1912 (when the Thiers fortifications were demilitarized).
Instead of a global transformation, most of these changes are concentrated on certain areas.

As illustrated in \Cref{fig:paris_changes}~(h), the modifications of outskirts are associated with infrastructure-oriented zones, particularly zones related to rail freight (in pink) and to waterborne freight and livestock activities (in blue). These areas experienced sustained change over multiple decades, reflecting continuous developments such as the construction of new tracks, facilities, and associated buildings.
In the meanwhile, other peripheral areas were spared from the extensive restructurings, as shown in yellow.
These relatively stable zones include former villages, such as Passy and Les Batignolles, that were gradually and organically integrated into the urban fabric, as well as newly planned districts such as Beau-Grenelle and parts of Belleville.

The modifications of central Paris
are also confined to specific sites that can be grouped into several categories, as shown in~\Cref{fig:paris_changes}~(i): 
hospitals (in yellow), parks (in green), railway stations (in pink), and a small number of other public buildings (e.g., prisons and military facilities, in blue). Additionally, areas shown in orange correspond to developments associated with the Paris World’s Fairs (Expositions Universelles) of 1878 (Trocadéro, Champ de Mars), 1889 (Eiffel Tower), and 1900 (Grand Palais, Petit Palais). The Louvre and the Jardin des Tuileries also underwent changes following the fire of the Palais des Tuileries during the Paris Commune in 1871 (not necessarily planned changes), as well as subsequent modifications to the garden layout.

\begin{figure}
    \centering
    \includegraphics[width=\textwidth]{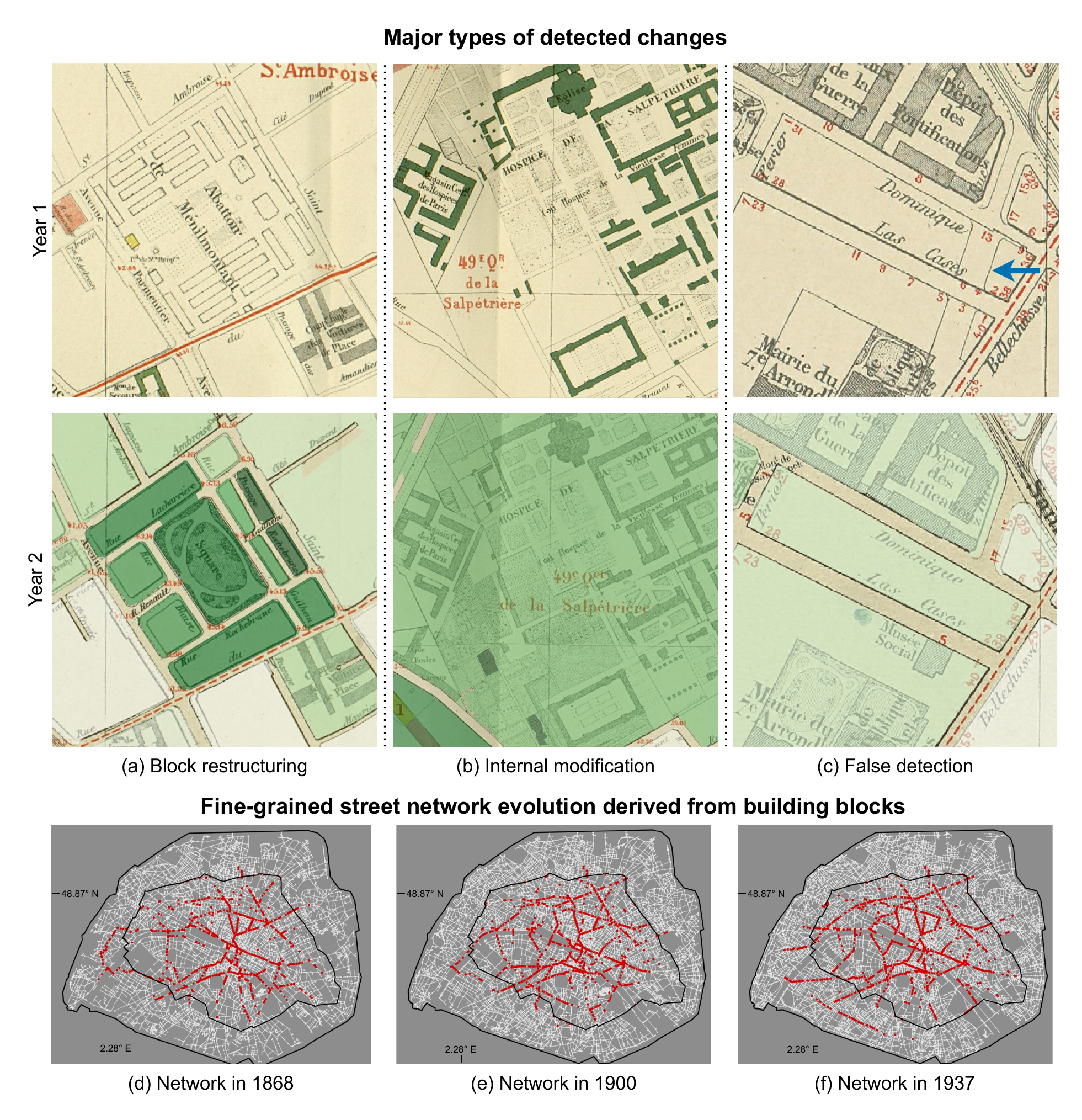}
    \caption{Major types of changed areas and additional applications of building block representations.
    Maps are sourced from~\citep{villeparis1868atlas, villeparis1878atlas, villeparis1900atlas, villeparis1912atlas}.
    Three main types of change are identified: (a) restructuring of building blocks driven by street openings and widenings; (b) infrastructure-related areas exhibiting internal modifications; and (c) apparent changes caused by false detections, typically due to construction lines indicating planned roads (blue arrow).
    Beyond change characterization, building block representations enable further analyses, such as street network evolution. We illustrate this by visualizing the betweenness centrality of intersections, where nodes in red correspond to the top 10\% most globally connected nodes.}
    \label{fig:types_of_change}
\end{figure}

Three main types of areas account for most of the detected change, as illustrated in~\Cref{fig:types_of_change}:
(1) areas that underwent major restructuring driven by street openings or street widenings, particularly within the Paris Zone (see (a));
(2) infrastructure-related areas with internal evolution, such as hospitals, transport and storage facilities, and parks, where localized additions or modifications alter block-level IoU values without large-scale reconfiguration (see (b)); and
(3) areas where the model fails to delineate building blocks accurately, typically due to construction lines indicating ongoing works or planned street projects (see (c)).

Overall, these results reveal a heterogeneous transformation geographically, with clusters of changes concentrated in recently annexed and infrastructure-oriented areas. The central area has a relatively stable spatial configuration, except for public infrastructures, parks, and major event-driven developments. 
The dominant forms of detected change correspond to street openings and widenings, as well as localized reconfigurations of large infrastructures. 
These data-driven spatial patterns are congruent with established interpretations of Parisian urban history~\citep{jones2006paris}, which describe the transformation of Paris during the third republic as a continuation of the syntax of urban improvement that characterized the urban transformation led by Haussmann during the Second French Empire---an urban framework whose structural legacy continued to shape the evolution of Paris well into the \nth{20} century, albeit at a noticeably slower pace.
We should also note that several major undertakings of the period---such as the expansion of the Métro through extensive underground works---are scarcely visible on the change maps. 
Additional applications of change profiling of Zurich can be found in Supplementary Note 7, which highlights the intensive industrial development and metropolitan growth during the late \nth{19} and early \nth{20} centuries.

In addition to the block-level change analysis, detailed building footprints and block boundaries provide a spatial framework for reconstructing street networks over time, enabling the analysis of their evolving structural roles in urban circulation. \Cref{fig:types_of_change}~(d-f) show the temporal evolution of betweenness centrality at street intersections in Paris, a network measure that captures their relative importance for connectivity by quantifying how frequently they lie on shortest paths between other locations~\citep{Freeman1977}. To emphasize the structural backbone of the street network, we display only locations with betweenness centrality values above the 90th percentile, corresponding to the top 10\% most globally connected intersections. The results show a progressive extension of highly central intersections from central Paris toward peripheral areas over time.

\subsection{Evaluation across map regimes}

We further evaluate the framework on Siegfried maps, the coarser topographic maps at 1:25{,}000, provided by the Federal Office of Topography swisstopo. The case-study results in Supplementary Note 7 show major urbanization processes in Zurich during the late \nth{19} and early \nth{20} century, including residential, industrial, and infrastructural expansion. As shown in Supplementary Figure 7, road-bounded blocks provide meaningful morphological units in dense urban areas where the road network is sufficiently dense. In peri-urban and suburban areas characterized by sparse infrastructure and discontinuous settlement patterns, the resulting units remain topologically valid but correspond less closely to coherent urban blocks.

We also evaluate the framework across map scales and cartographic series. In Supplementary Figure 8, we compare Old National Maps at 1:25{,}000 and 1:50{,}000 from the same time period (provided by swisstopo), thereby isolating differences in scale without introducing temporal change. The alignment module can normalize the positions of major structures across the two scales, whereas direct instance-level comparison becomes less reliable because the coarser representation generalizes and merges structures that remain distinct at the finer scale.

In Supplementary Figure 9, we evaluate alignment between Siegfried maps (1880s--1940s) and Old National Maps (1950s--1990s), both at 1:25{,}000. Despite differences in symbolization and style, the two series retain similar visual and spatial structures in roads, rivers, railways, and settlements, allowing the maps to be aligned. Their comparable scale and semantic granularity also allow the extracted instances to be compared across the two series.

We further assess sensitivity to inter-edition gaps in Supplementary Note 11 by grouping consecutive Siegfried map pairs according to their temporal separation. Larger temporal gaps are associated with greater alignment difficulty on average, while alignment remains feasible across the tested range when the compared editions retain sufficient common spatial and semantic structure.

\section{Discussion}

Detecting fine-grained change from historical maps is essential for many research fields but remains technically challenging due to severe spatial misalignment between map sheets and the often degraded graphic quality of earlier editions. Addressing these challenges requires both an alignment model capable of handling complex and non-uniform displacements, and an instance segmentation model that handles stylistic variation and visual degradation.  In this work, we address both alignment and instance extraction within a unified framework and, with both components performing favorably compared with the evaluated existing methods. Each component can also operate independently, allowing their use beyond the specific pipeline presented here.
We integrate these components into a comprehensive workflow that derives fine-grained urban change trajectories. 
The Paris case study demonstrates how these components operate together, with the resulting change representations further evaluated by historians and urban researchers. The experiments across different map regimes also illustrate how individual components can be adapted according to map scale, type, and urban content.

The proposed framework supports a transition  from ad hoc visual comparison to a quantitative, instance-based analysis of change in historical maps.
By using instance geometry as a proxy, the approach can accommodate visual discrepancies arising from differences in map production, printing, scanning, and storage conditions. 
The data-driven, bottom-up approach allows flexibility in how change can be conceptualized and analyzed, without imposing predefined structures.
The modular design additionally reduces the need for manual intervention at several stages of the workflow. Self-synthesis supports displacement estimation without manual correspondence annotation, while requiring vector annotations from only a limited subset of map sheets to improve robustness to temporal changes. Label propagation and temporal map fusion further reduce repeated vector annotation by reusing information across map editions. Together, these strategies make the workflow more practical for processing extended historical map sequences.

In our primary case study, we apply the framework to the large-scale urban map series Atlas Municipal at 1:5{,}000. Similar historical collections are also available. For example, the UK Ordnance Survey Town Plans and National Grid series~\cite{ordnance_survey_nls} from the \nth{19} to the \nth{20} century comprise more than 500{,}000 map sheets, with scales ranging from 1:1{,}056 to 1:10{,}000. Likewise, the Sanborn maps~\cite{sanborn_maps_loc} include more than 50{,}000 fire insurance maps, showing detailed building structures at scales such as 1:600, 1:1{,}200, and 1:2{,}400 over multiple dates in the \nth{20} century. These large-scale urban map series explicitly represent block interiors and exteriors, fine structural details, and often parcel subdivision, making them well suited for monitoring detailed intra-block change.

At coarser scales, the interpretation of resulting change profiles differs. For maps such as Siegfried at 1:25{,}000, the finest stable built units represented are typically building footprints or compact built-up masses. The resulting change profiles therefore reflect broader morphological changes rather than detailed internal block structure and can be interpreted in relation to processes such as urban consolidation, expansion, and restructuring. The suitability of block aggregation also varies spatially. In dense urban areas, road-bounded blocks can provide meaningful morphological units, whereas in peri-urban and suburban landscapes with sparse infrastructure and discontinuous settlement patterns, block-level change analysis becomes less informative. In such contexts, aggregation could instead incorporate parcel structure or other infrastructural boundaries, or changes could be interpreted directly at the building level.
Evaluation across map regimes indicates that comparable map scale and semantic granularity as well as sufficient visual and structural correspondences, whether within a single map series or across different but related series, are important conditions for applying the full framework.
Geometric alignment can remain feasible across different scales when sufficient common spatial structure is preserved, but meaningful instance-level change analysis requires greater consistency in scale and representation.

For historical urban analysis, fine-grained urban representations derived from historical maps, including building footprints, building blocks, and their associated change profiles, can provide structured reference data for examining the long-term evolution of urban form and tracing individual buildings in cities covered by large-scale historical mapping, such as London, New York, and Amsterdam.
In addition, detailed building footprints and block boundaries can provide a spatial basis for reconstructing street networks and examining changes in their structural roles over time.  
In the Paris case, the spread of highly central intersections toward peripheral areas suggests the development of alternative circulation routes outside the city center.
While similar patterns have been reported for the Paris street network~\citep{kirkley2018betweenness}, this example illustrates how fine-grained urban block representations can support investigations of the long-term street network evolution at a finer spatial scale across successive decades.
The extracted data can be further validated and corrected, enriched with additional attributes such as building names, and reused for analyses such as tracing block subdivision and consolidation or quantifying street openings and widenings.
By structuring geometric discrepancies at multiple spatial levels, from individual instances to aggregated building blocks, the framework can serve as a visual guide that highlights locations for closer inspection, helping researchers focus on areas of potential change or extraction and alignment errors rather than examining all map editions exhaustively.
When linked to other sources, such as professional activities, demographic data, or infrastructure investments, these representations may further support analyses of complex patterns of urban evolution and co-evolution dynamics between the built fabric and social or economic processes.

Beyond change detection within historical map series, the proposed alignment framework supports several related applications. The estimated displacement fields provide a quantitative description of geometric distortions between map editions and can be used to assess cartographic consistency, production practices, and spatial accuracy over time. Supplementary Figure 10 illustrates an example of map deformation over time in the Atlas Municipal series. Within a single map series, these fields enable longitudinal analysis of geometric distortions; across multiple series, they allow comparisons of cartographic reliability across periods or publishers.
The alignment results can also be reused for tasks such as label propagation and temporal data fusion. Although propagated labels may require correction because of object changes or residual alignment errors, they can provide a useful initialization for subsequent annotation. Temporal data fusion may be particularly useful for map sequences affected by local degradation, occlusion, or heterogeneous image quality. Together, these applications broaden the utility of the framework for the analysis and processing of historical cartographic datasets.

The framework has several limitations that point to directions for future work.
First, the applicability of the full framework is constrained by map scale, semantic granularity, visual and structural correspondence, and the availability of sufficient urban content.
In practice, this may limit the applicable data mainly to topographic and cadastral map series produced under relatively consistent cartographic standards, such as many systematic series from the \protect\nth{19} and \protect\nth{20} centuries. Besides, sufficient urban content is needed for meaningful block-level change profiling, which restricts the end-to-end applicability of the whole framework. 
Outside these scopes, individual components of the framework may still remain useful, but they require adaptation, or even higher-level adjustments, such as establishing semantic correspondences beyond pixel-based alignment for early maps characterized by strong abstraction, non-standardized symbolization or projection, and limited geometric consistency across years.

Second, the framework relies on the availability of digitized map scans, geo-referencing in a common coordinate system for initial coarse correspondence, systematic cataloguing for map retrieval, and suitable vector annotations. Although several established map archives exist, many historical maps have yet to be digitized and therefore require continued digitization and archiving efforts. Future work could improve digital archiving workflows and investigate methods for cases in which geo-referencing is missing or inaccurate, potentially by jointly optimizing geo-referencing and dense alignment within a unified framework.

Third, the current change representation should be understood as an IoU-based geometric proxy of structural persistence and potential change. In particular, a low IoU score may reflect genuine spatial reconfiguration, but it may also result from alignment and segmentation inaccuracies or other cartographic inconsistencies between the compared datasets. Therefore, the resulting change maps should be interpreted as indicators of geometric discrepancy rather than as fully validated records of historical change.
In addition, the IoU-based indicator intentionally treats all detections uniformly and does not distinguish between different types or levels of importance of change, such as demolition, reconstruction, functional conversion, or changes involving objects of varying relevance to urban studies. We leave such interpretations to historians and urban researchers, whose domain expertise is often required to contextualize historical transformations. Future work could support more specialized analyses by incorporating object categories, functional attributes, or graded change classes while preserving the flexibility of the current data-driven framework.

\section{Methods}\label{sec:methods}
\subsection{Data}
Our study utilizes historical urban planning maps from the Atlas municipal des vingt arrondissements de Paris~\cite{villeparis1868atlas, villeparis1878atlas, villeparis1888atlas, villeparis1900atlas, villeparis1912atlas, villeparis1925atlas, villeparis1937atlas}, produced at the scale of 1:5000. These maps depict buildings, building blocks, and other urban infrastructure of Paris and its suburbs. Accessed through the SoDuCO project~\cite{soduco}, the dataset consists of 16 individual map sheets and their subsequent editions (georeferenced), spanning 25 years between 1868 and 1937 and covering all 20 districts of Paris. 
Each map has approximately 10,000 $\times$ 7,400 pixels, with a spatial resolution of around 0.4 m/pixel. District masks are provided to crop each map, allowing a focused analysis of the central districts.
To ensure spatial consistency, we rectified the geo-referenced maps depicting the same locations by converting geographical coordinates to pixel coordinates with respect to a common origin, standardizing the resolution, and cropping to a uniform spatial extent.
For our analysis, we selected maps from the years 1868, 1878, 1888, 1900, 1912, 1925, and 1937 to monitor urban changes over time.

For displacement estimation, we have used the map covering the \nth{3}, \nth{4}, \nth{5}, \nth{6}, and \nth{7} districts for testing, while maps covering the remaining 15 districts are used for training. 
The maps were cropped into patches of 520 $\times$ 520 pixels using sliding windows, resulting in approximately 25,000 training pairs.
For building instance segmentation, we manually digitized two complete map sheets and propagated their vector labels to other editions using the estimated displacement fields. After post-processing, we obtained seven annotation maps in total. We trained on sequences of three consecutive maps, each cropped into smaller patches of 512 $\times$ 512 pixels, yielding about 4,000 sequences for training. 
For testing, we randomly selected a 2,000 $\times$ 2,000 pixel area from two different years in each of the remaining districts and annotated them using the same label-propagation strategy.

\subsection{Data synthesis}
For data synthesis, we synthesize object changes, text displacements, as well as displacement between maps. The details can be seen in Supplementary Figure 1.
\\
\textbf{Change synthesis.}
We randomly sample foreground objects based on the available vector annotations. Using a simple copy-and-paste method~\citep{dwibedi2017cut}, we generate a modified copy from the original image under four scenarios: ``addition'' if the foreground objects are pasted into the copy, ``deletion'' if the foreground objects are pasted into the original image, ``replacement'' if both addition and deletion happen, and ``unchanged'' if no foreground objects are used. This setup enables the model to learn from diverse types of object changes.
\\
\textbf{Text-displacement synthesis.}
To extract textual elements, we borrow the off-the-shelf state-of-art text detection model DeepSolo~\citep{Ye2023deepsolo} based on the transformer architecture. 
We fine-tune it on the open dataset ICDAR 2024 Competition for Historical Map Text Detection, Recognition, and Linking~\citep{li2024icdar}.
Similarly, we use the copy-and-paste strategy to simulate random text displacements on both the original image and its copy, and integrate these perturbations with the change synthesis process.
\\
\textbf{Displacement synthesis.}
To simulate realistic geometric distortions, we apply a range of transformations\textemdash including homography, affine, thin plate splines (TPS), and their combinations~\citep{rocco2017geometric}\textemdash to the copied image. The corresponding ground-truth displacement fields are then used to supervise the model. To increase robustness to appearance variations, we randomly adjust brightness, saturation, and contrast while keeping hue unchanged, as hue often encodes semantic class information in maps (e.g., hydrological features are typically blue). We also apply subtle Gaussian blurs to introduce minor texture variations, further enhancing the model’s resilience to visual inconsistencies.

\subsection{Displacement estimation}
\textbf{Problem definition.} 
Given two map patches $I \in \mathbb{R}^{H \times W \times 3}$ and $J \in \mathbb{R}^{H \times W \times 3}$ depicting the same area of the same size~$H\times W$, our goal is to estimate pixel-wise displacements that are pixel-wise 
2D motion vectors, $F_{I \rightarrow J} \in \mathbb{R}^{H \times W \times 2}$, often also denoted as optical flow, in the coordinate system of image $I$, that relates pixels in $I$ to $J$.
\\
\textbf{Triplet consistency.}
Given a real image pair $I,J$, we construct an image triplet comprised of $I$,$J$, and $I'$, with $I'$ being a synthesized copy of $I$ augmented with synthesized changes, randomly displaced texts, and a known displacement field $W$.
The displacement consistency along the triplet between W and the composition from $I' \rightarrow J \rightarrow I$ can be modeled:
\begin{equation}
    W = \widehat{F}_{I'\rightarrow J} + {\phi}_{\widehat{F}_{I'\rightarrow J}}(\widehat{F}_{J\rightarrow I}), 
\label{eq:consistency_flow}
\end{equation}
where ${\widehat{F}_{I'\rightarrow J}}$ is the estimated displacement from $I'$ to $J$ and ${\widehat{F}_{J\rightarrow I}}$ is the estimated displacement from $J$ to $I$. 
${\phi}_{\widehat{F}_{I'\rightarrow J}}$ warps the $\widehat{F}_{J\rightarrow I}$ in the coordinate system of $J$ into that of $I'$, to align with $\widehat{F}_{I'\rightarrow J}$.
This consistency is used to refine the displacement estimation ${\widehat{F}_{I'\rightarrow J}}$ and ${\widehat{F}_{J\rightarrow I}}$.
The consistency is enforced only in valid regions where pixels from both sides represent a valid, non-occluded mapping and do not fall outside the image boundaries~\citep{truong2021warp}.
\\
\textbf{Network structure.}
Given the triplet formed by $I$, $J$ and the modified copy $I'$, following WarpC~\citep{truong2021warp}, we use GLU-Net~\citep{truong2020glu} for displacement estimation between each pair ${\widehat{F}_{I'\rightarrow J}}$, ${\widehat{F}_{J\rightarrow I}}$ and ${\widehat{F}_{I'\rightarrow I}}$ with shared parameters $\theta$. The architecture of GLU-Net can be found in Supplementary Figure 2. It estimates the displacement fields through a hierarchical workflow, which can capture and align multi-scale details. The feature extractor extracts four levels of image features in total.  At each feature level, the correlation between the extracted features from the paired inputs is computed and used to estimate the displacement field at this resolution.  
\\
\textbf{Loss function.} 
In the constructed triplet ($I,J,I'$), the ground-truth displacement $W$ between $I'$ and $I$ is known at the original resolution $H \times W$. We can obtain the direct supervision loss:
\begin{equation}
    \mathcal{L}_{I'\rightarrow I} = \sum_l \alpha^l||\widehat{F}_{I'\rightarrow I}^l - W^l||,
    \label{eq:supervise}
\end{equation}
where $l$ indicates the feature level and $\alpha^l$ is the scalar factor to weight the loss at this level. $W$ is down-sampled it to $H_l \times W_l$ each level $l$ to calculate the multi-level loss.
According to the equation \eqref{eq:consistency_flow}, we can enforce the consistency between the composition flow $I \rightarrow J \rightarrow I'$ and the known flow $I' \rightarrow I$ at multiple levels:
\begin{equation}
    \mathcal{L}_{I'\rightarrow J \rightarrow I} = \sum_l \alpha^l ||\widehat{F}_{I'\rightarrow J}^l + {\phi}_{\widehat{F}_{I'\rightarrow J}^l}(\widehat{F}_{J\rightarrow I}^l) - W^l||.
    \label{eq:warp_c}
\end{equation}
By combining equations \eqref{eq:supervise} and \eqref{eq:warp_c}, we have the overall objective:
\begin{equation}
    \mathcal{L} = \sum_{(I,I')}\mathcal{L}_{I'\rightarrow I} + \sum_{(I,I',J)}\mathcal{L}_{I'\rightarrow J \rightarrow I}.
    \label{eq:overall_obj}
\end{equation}

\subsection{Instance extraction}
In our work, we adapt Mask2Former~\citep{cheng2021mask2former} for segmenting building instances, a state-of-the-art architecture suitable for both semantic and instance segmentation based on the transformer architecture~\citep{vaswani2017attention}.
Compared with CNNs, transformers handle long-range context by modeling the global relationship between all locations based on the self-attention mechanism~\citep{vaswani2017attention} and have shown remarkable improvements in image segmentation~\citep{cheng2021maskformer, cheng2021mask2former}. 
We argue that the long-range context is especially advantageous for segmenting our maps, where most building instances are represented as polygons without any textures. 
Rather than segmentation per pixel, Mask2Former uses mask classification that groups pixels into $N$ segments as $N$ binary masks. Each mask is assigned both an instance label and a corresponding class category label.
These masks guide the transformer's attention to focus on features within the masked region, thereby improving training efficiency.
In our case, we define two classes\textemdash roads and building units, the latter defined as closed-shape polygons. The model outputs a predicted class and instance ID for each pixel.

To overcome degrading graphic quality, we add multi-scale feature fusion blocks given features from different years. Our modified architecture is illustrated in Supplementary Figure 3. Following the original architecture, given a sequence of images, our model has a shared backbone to extract image features, a shared pixel decoder to restore the original image resolution from the backbone features, and a transformer decoder that predicts the instance mask and class from the multi-level pixel features. To fuse the features from different images at each feature level, we use the cross-attention module, followed by a Feed Forward Network (FFN), which is comprised of two linear transformations with ReLU activation and dropouts in between. Note that we apply the estimated displacement fields to spatially align the map 
editions beforehand to ensure better feature correspondences at the same location across years. 
At inference, the maps are cropped into tiles of 512 $\times$ 512 pixels with 50\% overlap, and the model generates predictions for each tile.
To reconstruct complete instances across the entire map sheet, we identify the same instance across overlapping tiles and unify the instance ID. The corresponding pseudo-code is presented below.
For each tile, instances are compared with those in the overlapping regions of the adjacent tiles---specifically, the tile directly above and the tile to the right. The Intersection over Union (IoU) is calculated for each pair of instances. The ID of the overlapping instance with the highest IoU (greater than 0) is assigned to the current instance.
To prevent duplicated assignments, this process is performed for either the tile above or the tile to the right, but not both simultaneously.

\begin{algorithm}[H]
\caption{Stitching Instance Predictions across Overlapping Tiles}
\label{alg:stitch}

Input: patch predictions $\mathcal{P}$, patch size $s$, step size $t$, overlap $o$, image height $h$, image width $w$, IoU threshold $\tau$;

Output: stitched instance image $I$, where background pixels are labeled as $0$, and instance pixels are labeled with positive IDs.

Initialize: output image $I$ with all pixels set to $-1$, rows $r \leftarrow h/t - 1$, cols $c \leftarrow w/t - 1$, global instance ID $g \leftarrow 1$;

\For{$i$ in $r$}{
    \For{$j$ in $c$}{
        Extract the current patch 
        $p \leftarrow \mathcal{P}[i * c + j]$

        Identify all instance IDs in the patch, excluding background:
        $\mathcal{L} \leftarrow \{\ell \mid \ell \in p\}$

        \For{$\ell$ in $\mathcal{L}$}{
            Extract the mask of the current local instance:
            $M \leftarrow (p = \ell)$

            Initialize matched IDs:
            $id^u \leftarrow \text{None}$, $id^l \leftarrow \text{None}$

            \If{there is an upper tile}{
                Extract the upper overlap region:
                $R^u \leftarrow I[i*t:i*t+o,\ j*t:j*t+s]$

                Find the best matching instance $id^u$ in $R^u$ with the highest IoU with $M[:o,:]$, when IoU $> \tau$
            }

            \If{there is a left tile}{
                Extract the left overlap region:
                $R^l \leftarrow I[i*t:i*t+s,\ j*t:j*t+o]$

                Find the best matching instance $id^l$ in $R^l$ with the highest IoU with $M[:,:o]$, when IoU $> \tau$
            }

            \If{$id^u \neq \text{None}$ and $id^l \neq \text{None}$ and $id^u \neq id^l$}{
                Merge the two matched IDs:
                $id \leftarrow \min(id^u, id^l)$

                Replace the larger ID in $I$ with $id$
            }
            \ElseIf{$id^u \neq \text{None}$}{
                Assign the upper matched ID:
                $id \leftarrow id^u$
            }
            \ElseIf{$id^l \neq \text{None}$}{
                Assign the left matched ID:
                $id \leftarrow id^l$
            }
            \Else{
                Assign a new global ID:
                $id \leftarrow g$

                $g \leftarrow g + 1$
            }
            Assign $id$ to the current instance region in
            $I[i*t:i*t+s,\ j*t:j*t+s]$, without overwriting pixels already assigned to another instance.
        }
        Fill all remaining unwritten pixels in $I[i*t:i*t+s,\ j*t:j*t+s]$ with background value 0.
    }
}

\Return{I}
\end{algorithm}

\subsection{Evaluation metrics}
\textbf{Displacement smoothness.}
We measure the smoothness of the displacement field using mean variation (mV). It calculates the average of the absolute gradient between neighboring pixels:
\begin{equation}
\text{mV} = \frac{1}{N} \sum_{(x, y) \in \Omega} \sum_{(i,j) \in \mathcal{N}} \left\| \mathbf{u}(x, y) - \mathbf{u}(x+i, y+j) \right\|_1,
\end{equation}
where $ \mathbf{u}(x, y) $ is a displacement vector at the position $(x,y)$.  $N$ is the number of pixels, and $\mathcal{N}$ is a set of neighbor offsets, e.g., $\{(1,0), (0,1), (-1,0), 0,-1)\})$ for 4-neighborhood.
\\
\textbf{Displacement consistency.}
Similar to~\Cref{eq:warp_c}, we evaluate the L1 consistency of the displacement estimation within a triplet from three consecutive years $T_1, T_2, T_3$:
\begin{equation}
    L1 = ||\widehat{F}_{T_1\rightarrow T_2} + {\phi}_{\widehat{F}_{T_1\rightarrow T2}}(\widehat{F}_{T_2\rightarrow T_3}) - \widehat{F}_{T_1\rightarrow T_3}||.
    \label{eq:l1_consistency}
\end{equation}
\\
\textbf{Image-level comparison.}
We use the the structural similarity index (SSIM) to measure the image similarity:
\begin{equation}
    SSIM = \frac{(2\mu_x\mu_y + c_1)(2\sigma_{xy}+c_2)}{({\mu_x}^2+{\mu_y}^2+c_1)({\sigma_x}^2+{\sigma_y}^2+c_2)},
\end{equation}
where $\mu_x$ and $\mu_y$ are average values; $\sigma_x$ and $\sigma_y$ are variances; $\sigma_{xy}$ is the covariance of x and y; $c_1 =(k_1L)^2$ and $c_2 =(k_2L)^2$ are constants with $k_1=0.01, k_2=0.03$ and L being the dynamic range of the image pixel values. 
For RGB images, the per-channel RMSE and SSIM are calculated and averaged.
\\
\textbf{Object-level comparison.}
We use Chamfer distance (CD) to measure the average shortest distance between objects from two sets. CD between the two point sets $S_1$ and $S_2$ is defined as:
\begin{equation}
    CD = \frac{1}{|S_1|}\sum_{x \in S_1} {min}_{y \in S_2} \|x-y\|^2 +\frac{1}{|S_2|}\sum_{y \in S_2} {min}_{x \in S_1} \|y-x\|^2.
\end{equation}
Each point $x \in S_1$ finds its closest neighbor in $S_2$ and vice versa.  All point-wise distances are averaged. In our case, CD measures the distance between the sets of vertices filtered by the specific distance threshold.
\\
\textbf{Instance-level comparison.}
The average precision (AP)~\citep{lin2014microsoft} score is normally used to evaluate the accuracy for instance segmentation. It's based on the precision and recall scores. 
The Hungarian matching~\citep{carion2020detr} is used to assign a one-to-one matching between predicted instances and the ground truth as a standard approach. 
A predicted instance is considered correct if its Intersection over Union (IoU) with the matched ground-truth instance exceeds a predefined threshold $\theta$.
Precision measures how many of the predicted instances are correct, while recall measures how many of the ground-truth instances are correctly detected.

\begin{equation}
    Precision = \frac{TP}{TP+FP},
\end{equation}
\begin{equation}
    Recall = \frac{TP}{TP+FN},
\end{equation}
And the AP is the area under the precision-recall curve (AUC):
\begin{equation}
    AP = \int_{0}^{1}p(r)dr,
\end{equation}
where $p$ is the precision at recall $r$, computed at a certain IoU threshold $\theta$, typically larger than 0.5.
In practice, this is calculated using discrete values by averaging precision across all recall levels~\citep{pascalvoc2012, coco2019detection, objectdetection2021}.

\newpage
\section*{Data availability}
The block-level change profiles generated in this study have been deposited in \url{https://doi.org/10.6084/m9.figshare.33386371}.
The map images were openly accessible through the SoDuCo project (\url{https://soduco.geohistoricaldata.org}) via the GeoHistoricalData IIIF server (\url{https://iiif.geohistoricaldata.org}), together with associated metadata and georeferencing annotations. The original map images were provided to the SoDuCo project by the Bibliothèque de l'Hôtel de Ville (BHdV), Ville de Paris, with permission for redistribution subject to appropriate attribution.
The annotations for instance segmentation can be accessed here \url{https://doi.org/10.6084/m9.figshare.33386371}. The source data for Figure 4 are provided with this paper.

\section*{Code availability}
All code used in this study, including modules for displacement estimation, instance extraction, instance stitching, change quantification, and object geo-referencing, can be accessed via \url{https://github.com/sian-wusidi/map_alignment}. The model weights for displacement estimation and instance segmentation are also provided.

\newpage
% \renewcommand{\contentsname}{List of supplementary items}
% \tableofcontents
\begin{appendices}
\setcounter{figure}{0}
\renewcommand{\figurename}{Supplementary Fig.}
\renewcommand{\tablename}{Supplementary Tab.}
\setcounter{table}{0}
%\renewcommand{\thetable}{S\arabic{table}}
% Change cleveref reference names
\crefname{figure}{Supplementary Fig.}{Supplementary Figs.}
\Crefname{figure}{Supplementary Fig.}{Supplementary Figs.}

\crefname{table}{Supplementary Tab.}{Supplementary Tabs.}
\Crefname{table}{Supplementary Tab.}{Supplementary Tabs.}

\newpage
\section*{Supplementary Note 1: Technical terms}
\addcontentsline{toc}{section}{Supplementary Note 1: Technical terms}

\textbf{Geo-referencing. }
Geo-referencing is the process of associating spatial data (such as maps, aerial images, or satellite photos) with specific geographic coordinates in a particular coordinate system. 
\\
\\
\textbf{Image rectification. }
Image rectification refers to applying a geometric transformation to align images to a common reference frame. In this paper, it specifically denotes transforming images so that they share the same origin, spatial resolution, and image size.
\\
\\
\textbf{Image matching.} Image matching is a fundamental computer vision problem that aims to find correspondences between image pairs.
\\
\\
\textbf{Map cropping}. Map cropping is the process of partitioning a large map sheet, which is often too large to process directly, into smaller patches such as 256×256 pixels.
\\
\\
\textbf{Displacement estimation.} Displacement estimation is the process of estimating a pixel-wise displacement field between two images by computing the shift vectors of corresponding pixels, enabling image alignment.
\\
\\
\textbf{Warping.} Warping is the process of geometrically transforming a source image by shifting its pixels according to a displacement field so that they align with corresponding positions in a target image.
\\
\\
\textbf{Vector annotation.} Vector annotation refers to the process of manually or semi-automatically labeling map features such as buildings, roads, or rivers, by annotating them as vector geometries such as points, lines, or polygons.
\\
\\
\textbf{Label propagation.} Label propagation refers to assigning labels to unlabeled pixels or regions by propagating known labels through spatial or temporal relationships. In this paper, it refers to propagating labels across maps from different years.
\\
\\
\textbf{Instance segmentation.} Instance segmentation is the task of simultaneously performing object detection and pixel-wise segmentation, such that each object instance is segmented and assigned a unique instance ID.
\\
\\
\textbf{Cross attention.} Cross attention is an attention mechanism where the query and the key–value pairs are derived from different feature sets, enabling one representation to selectively attend to and integrate information from another.

%\\
%\textbf{Convolutional Neural Networks (CNNs). }
%\textbf{Transformers}

\newpage

\section*{Supplementary Note 2: More details of methodology}
\addcontentsline{toc}{section}{Supplementary Note 2: More details of methodology}
Details of triplet cycle consistency, data synthesis, the displacement estimation network and instance segmentation network.

\begin{figure}[H]
    \centering
    \includegraphics[width=\linewidth]{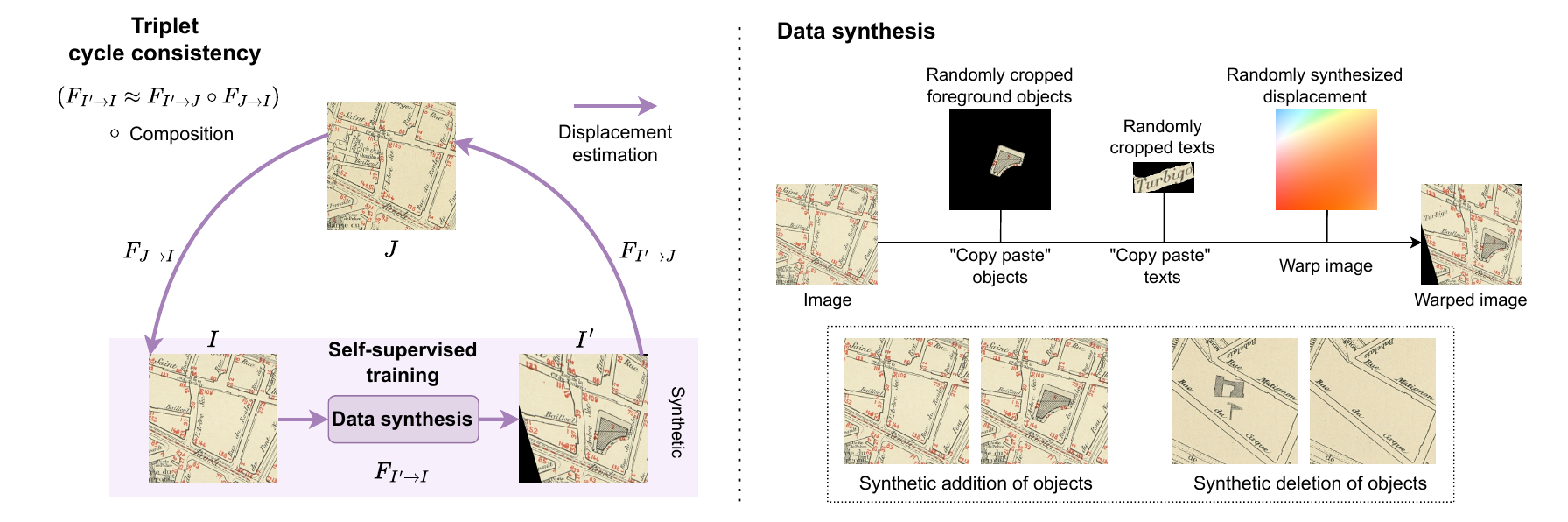}
    \caption{Detailed illustration of triplet cycle consistency and self-supervised data synthesis used for displacement estimation. Images are synthesized with known displacements, with random text and object injections to train the network to ignore these disturbances.}
    \label{fig:triplet_and_synthesis}
\end{figure}

\begin{figure}[H]
    \centering
    \includegraphics[width=0.76\linewidth]{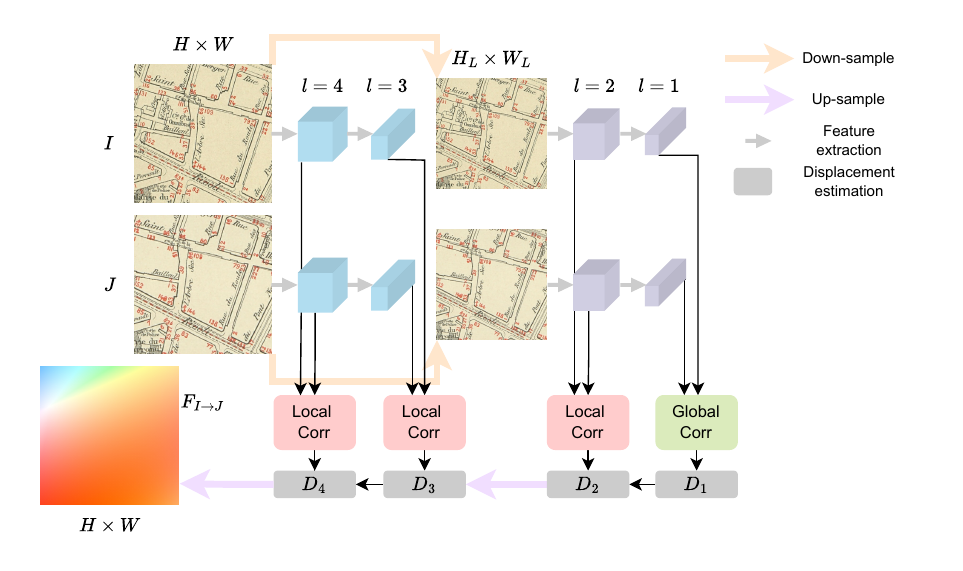}
    \caption{For displacement estimation, we use the GLU-Net architecture~\citep{truong2020glu}. The paired images $I$, $J$ are first down-sampled. Both original and down-sampled images are fed into feature extractors, resulting in multi-level features. Local and global correlations between the source and target features are computed at multiple levels $l \in {1,2,3,4}$. Using these correlations, 
    multi-level displacement fields are iteratively estimated by refining the displacement from the previous level with the current correlation information. The estimated displacement field from the final layer is up-sampled to produce the final estimation. }
    \label{fig:glunet}
\end{figure}

\begin{figure}[H]
    \centering
    \includegraphics[width=0.9\linewidth]{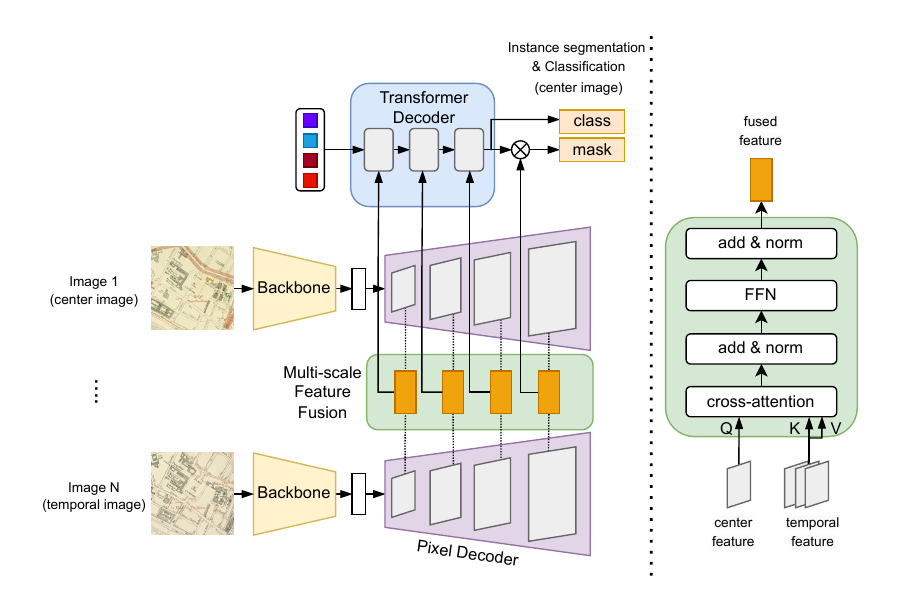}
    \caption{Our proposed architecture for instance segmentation. We extend the Mask2former model by introducing multi-scale feature fusion blocks to integrate features from maps at different years, based on the cross-attention mechanism. Note that we apply the estimated displacement fields to spatially align the map editions across years beforehand. }
    \label{fig:mask2former_modify}
\end{figure}

\section*{Supplementary Note 3: Experiment details}
\addcontentsline{toc}{section}{Supplementary Note 3: Experiment details}

For displacement estimation, we trained our model with a batch size of 12 on NVIDIA RTX 3090 GPU. We used the Adam Optimizer with a learning rate of 1e-4 and a weight decay of 4e-4. Given the substantial size of our dataset (25,000 pairs), the model was trained for 20 epochs, where convergence was observed.
For instance segmentation, we trained our model with a batch size of 4 on NVIDIA RTX 4090 GPU. We used the Adam optimizer with a learning rate of 5e-5 and we trained the model for 25 epochs, given the dataset of 4,000 sequences, each consisting 3 images.

\section*{Supplementary Note 4: Related work}
\addcontentsline{toc}{section}{Supplementary Note 4: Related work}

\subsection*{Image matching}
\label{sec:related_work_image_matching}
Image matching can be classified into sparse and dense matching. 
Sparse matching methods rely on detecting stable, representative, repeatable, and sparse key points across images~\citep{lowe2004distinctive, bay2006surf, rublee2011orb, truong2019glampoints, yi2016lift, detone2018superpoint, ono2018lf}. Typically, local image features from key points are matched to estimate the geometric transformation (e.g. affine or homography) between images~\citep{szeliski2007image}.
On the other hand, dense matching estimates the per-pixel correspondence and displacement between images, especially with the help of deep learning methods. 
It has mostly been addressed in the context of optical flow~\citep{baker2011database, hur2020optical, shen2020ransac, dosovitskiy2015flownet, sun2018pwc, sun2019models, hui2018liteflownet} or geometric matching~\citep{melekhov2019dgc, truong2020glu, truong2023pdc, truong2021warp, shen2020ransac, park2022dual}, where image pairs are from consecutive frames in a video or from different views of the same scene. 

Dense matching methods can be either supervised or unsupervised.
Supervised methods require a carefully designed synthetic dataset with ground-truth flow/displacement annotations~\citep{dosovitskiy2015flownet, mayer2016large}.
Unsupervised methods, on the other hand, do not require ground-truth annotations. Instead, they often use photometric consistency to encourage the estimated flow or displacement to align locations with similar appearances~\citep{ren2017unsupervised, shen2020ransac, jonschkowski2020matters}. At the same time, the spatial smoothness of the estimated flow or displacement is typically assumed and enforced.
For example, RANSAC-Flow~\citep{shen2020ransac} uses the photogrammetric loss combined with the cycle consistency between the forward and backward flow estimation. It combines both sparse and dense matching methods by using sparse features to pre-align images with homography and then estimates the per-pixel dense optical flow using neural networks.
Another approach of unsupervised dense matching synthesizes warps of real images by combining different geometric transformations and uses the synthesized flow/displacement for self-supervision~\citep{truong2020glu, truong2023pdc, park2022dual, melekhov2019dgc}.
Those methods can handle larger displacements by adjusting the amplitude of the synthesized warp and are robust against appearance changes without relying on photometric losses.
GLU-Net~\citep{truong2020glu} integrates both global and local feature correlations to handle large and small displacements.
Direct supervision trains the network to regress a randomly sampled displacement field, but this often results in poor generalization to real image pairs. 
To solve this problem, WarpC-GLU-Net~\citep{truong2021warp} introduces an image triplet strategy, combining real image pairs with a randomly warped version of one image and enforcing consistency within the triplet.
Starting from this method, we add map-specific disturbances, i.e. changes and randomly displaced texts, to enhance the displacement estimation for aligning historical maps.

% %
% Unlike change detection methods from image analysis, which require exact spatial co-incidence of objects across time, our approach accounts for misalignments commonly found in historical maps. And unlike existing work about historical map analysis, our work goes beyond geo-referencing~\cite{Jonas2021georef, Jonas2021content, Varela2018georef}: per-pixel displacement fields are reconstructed to align maps and to propagate annotations across time. Moreover, we also account for differences in the placement of text and figures, contrary to~\cite{wu2022unsupervised}. 
\subsection*{Historical map registration and alignment}
In existing research, map registration is often treated as a step within the geo-referencing process, where coordinates are assigned to maps by comparing them with reference maps or other geospatial data, typically using control points~\citep{barazzetti2014historical, vaienti2025exploring, Vaienti2025Georeferencing}, image features~\citep{deng2024feature, zambanini2019feature}, extracted objects~\citep{Jonas2021georef, Jonas2021content, yan2017polygon}. 
\cite{sun2021aligning} introduced a geographic knowledge graph (GKG), an organized collection of entities and concepts interconnected through their relationships, allowing the identification of the same object at different years. However, their pipeline required control points to perform subsequent geometric transformations.
Although these methods are effective, they typically depend on a single pre-defined or explicitly identified geometric transformation. 
In a novel approach, \cite{wu2022unsupervised} employed deep neural networks to directly establish dense correspondences between maps, which can handle non-geometric deformation and is fully automated without defining the transformations and control points.
Yet, this approach remains sensitive to object changes and text displacements, which can introduce significant distortions in the alignment process.

\subsection*{Historical map segmentation}
Unlocking the information from historical maps offers valuable insights into past landscapes and the spatio-temporal evolution of Earth’s features. Processing these maps has become an interdisciplinary effort between cartography, geoinformatics, and computer vision. In contrast to methods that rely on manual interventions~\citep{bin1998system, dhar2006extraction, leyk2010segmentation} or map-specific fine-tuning of certain features~\citep{khotanzad2003contour, samet2012new, chiang2013efficient}, deep learning has facilitated the automated segmentation of historical maps. This includes detecting roads~\citep{ekim2021road, rath2023settlement, jiao2024extracting}, buildings and blocks~\citep{Uhl2020building, Heitzler2020building, chen2021combining, rath2023settlement}, water bodies~\citep{wu2022leveraging, wu2022closer, o2024wetland}, and text~\citep{li2021synthetic, lin2024hyper}, using CNNs and transformers.
While these works mainly address texture-rich maps, ~\cite{chen2021combining, chen2021vectorization, chen2024benchmark} focused on segmenting building instances from texture-less maps, specifically the \textit{Atlas Municipal}, by conducting edge detection first and consequently converting the detected edges to closed polygons that form building instances. 
However, these edge-based methods fail to detect sidewalks\textemdash key elements of urban infrastructure\textemdash typically represented with open borders in maps. 

%\section{Additional technical details}
\section*{Supplementary Note 5: Qualitative comparison of estimated displacement field}
\addcontentsline{toc}{section}{Supplementary Note 5: Qualitative comparison of estimated displacement field}
\begin{figure}[H]
    \centering
    \begin{tabularx}{\linewidth}{cc}
    \includegraphics[width=0.2\linewidth]{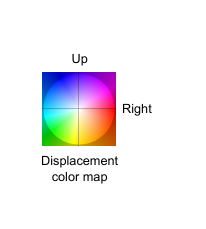} &  \includegraphics[width=0.56\linewidth]{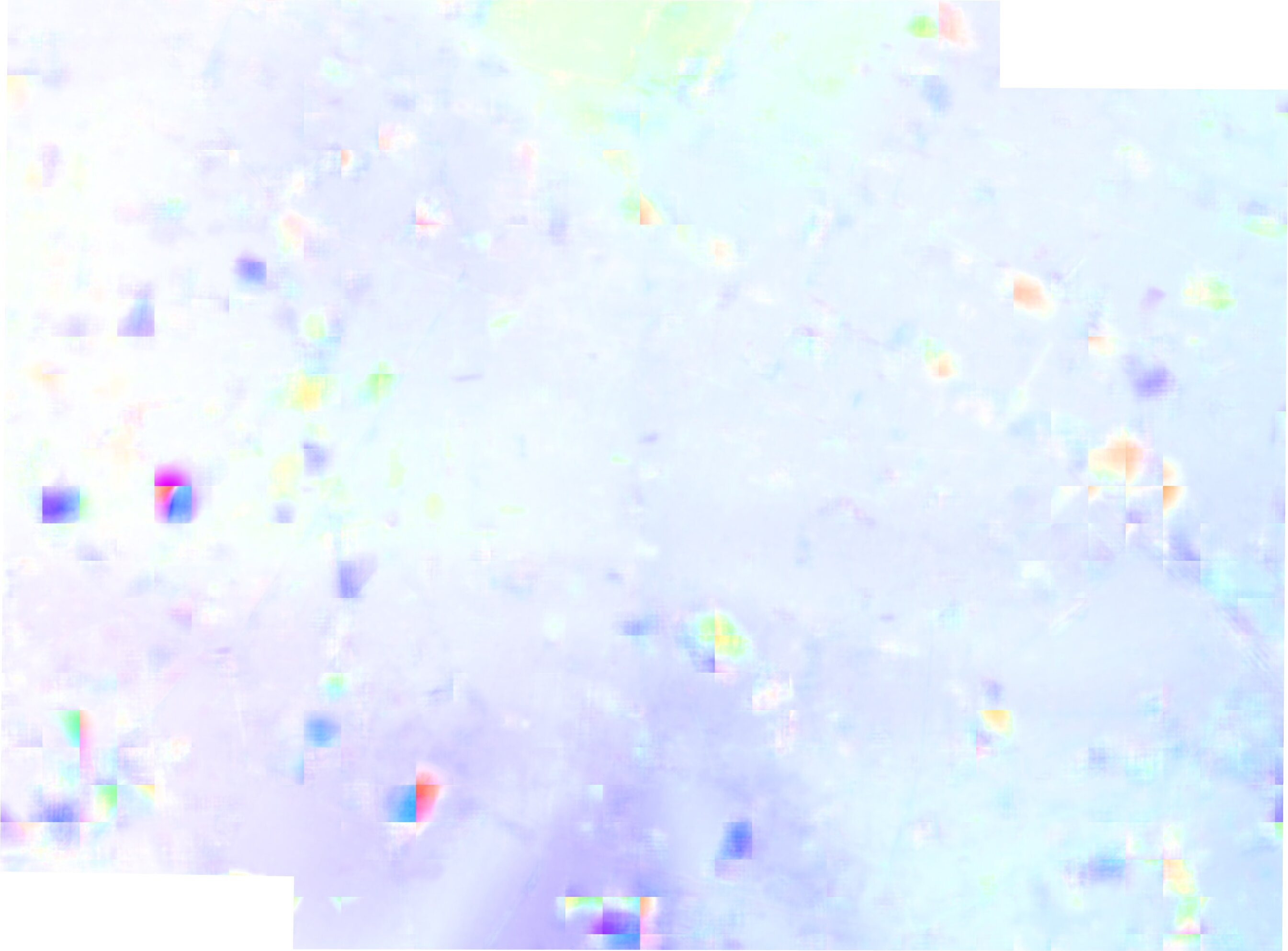}\\
    & \includegraphics[width=0.56\linewidth]{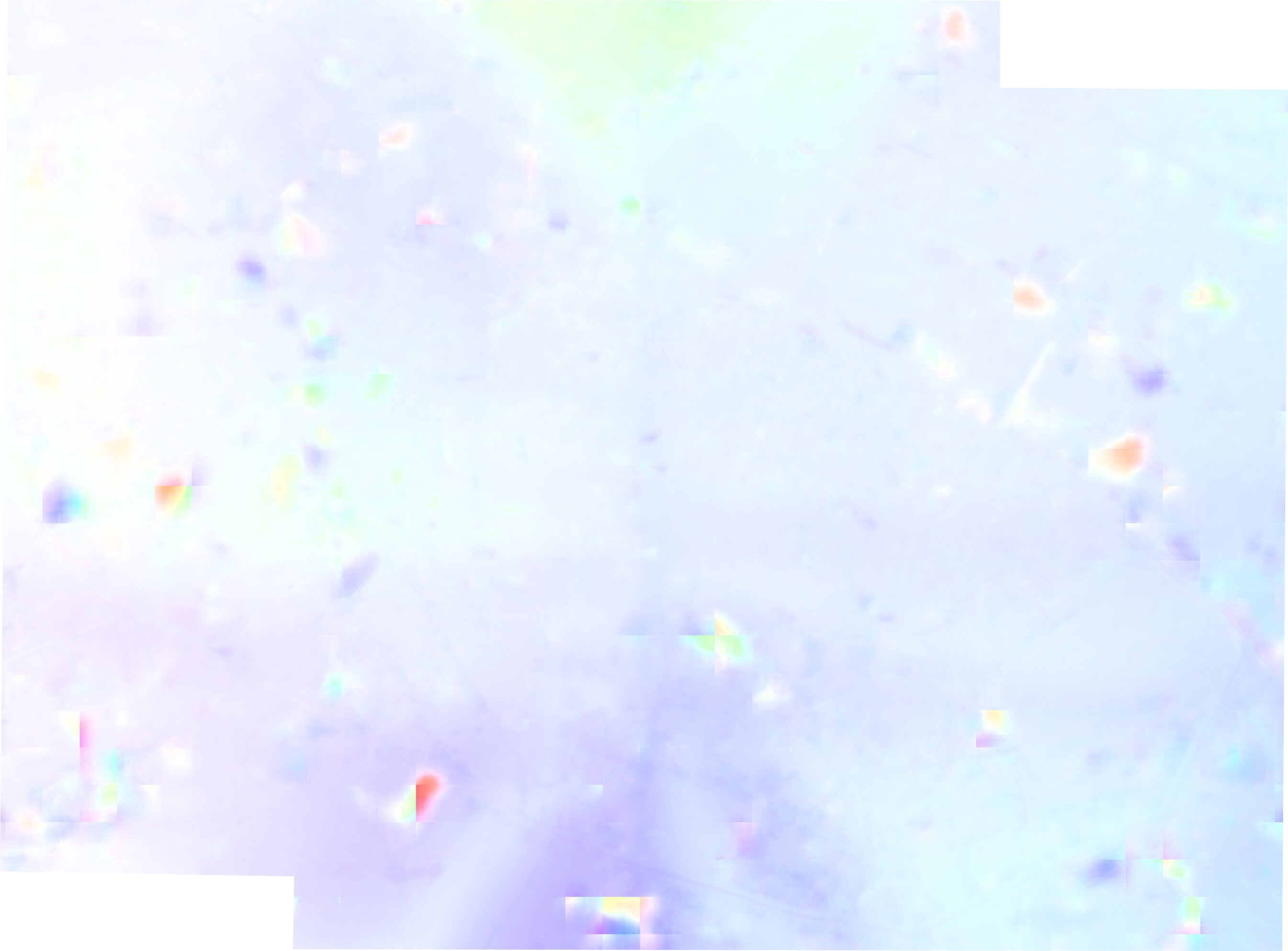}\\
    & \includegraphics[width=0.56\linewidth]{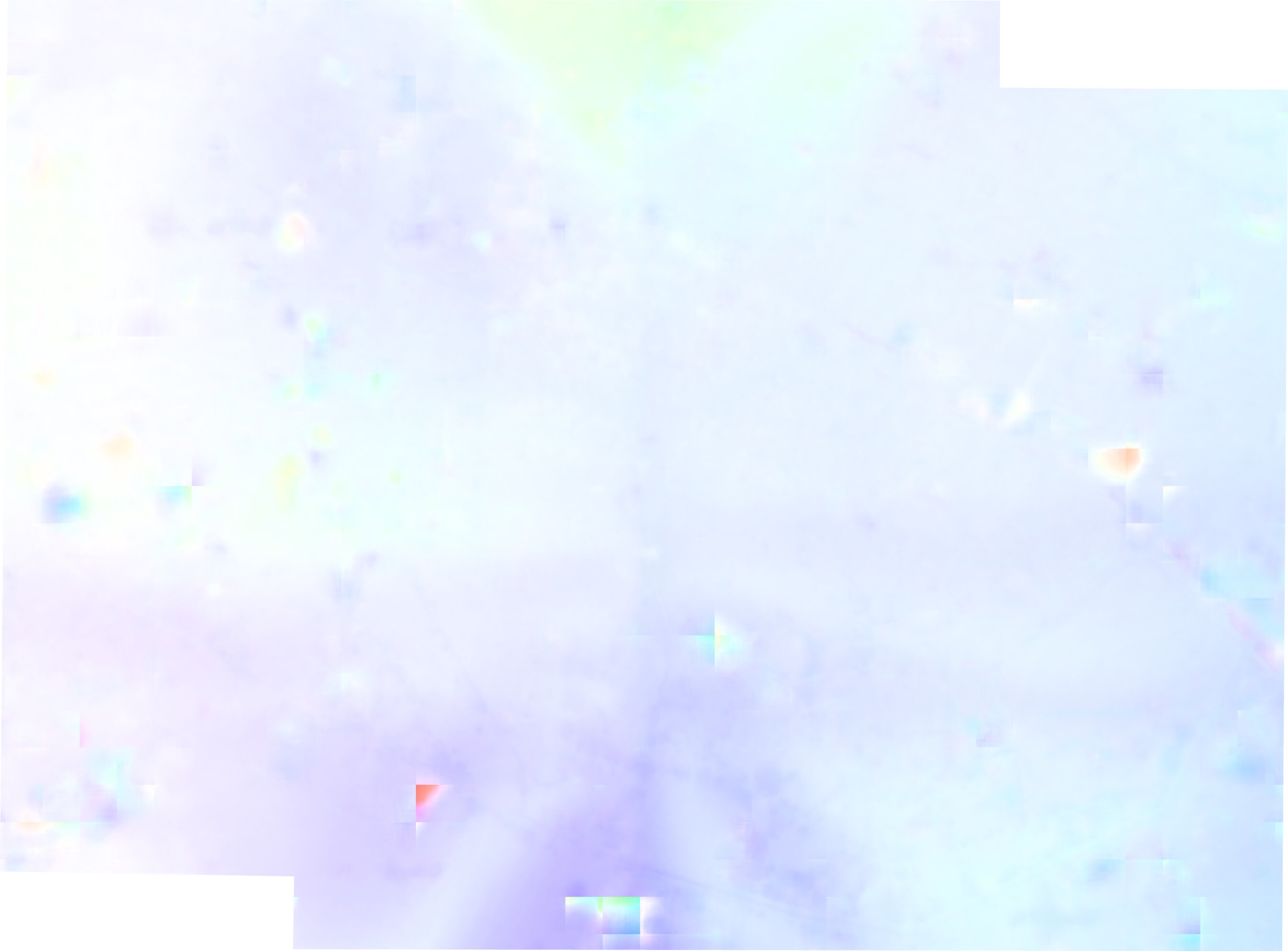}\\    
    \end{tabularx}
    \caption{A qualitative comparison of the estimated displacement field tested on maps of the same area from the years 1905 and 1937. The results, shown from top to bottom, include GluNet, WarpC, and our method. Our model outperforms the others by generating a smoother displacement field and effectively revealing the upward-right shift and the paper-folding artifacts. } % \url{(https://iiif.geohistoricaldata.org/iiif/3/BHdV_PL_ATL20Ardt_1937_0003.jpg/)
    \label{fig:quali_flow}
\end{figure}

\section*{Supplementary Note 6: Additional benchmark on other maps}
\addcontentsline{toc}{section}{Supplementary Note 6: Additional benchmark on other maps}
We re-trained our displacement estimation model on Siegfried maps, the historical topographic map series at the scale of 1:25,000, provided by the Federal Office of Topography swisstopo~\cite{siegfried_map}. For training, we used 5 map sheets, each at 5-6 time steps of the size 7000 $\times$ 4800. The maps were cropped into patches of 520 × 520 pixels using sliding windows, resulting in approximately 12,000 training pairs. To simulate foreground object changes and text displacements, we use building vector annotations from three map sheets and text detection results from two map sheets, respectively. For testing, we used three additional map sheets, each at 3-6 time steps. We calculate the chamfer distance (CD) of buildings. We masked out landscape features like rivers and lakes in calculating total variation and temporal consistency as they are irrelevant for urban analysis. As shown in~\Cref{tab:additional_benchmark_alignment}, our model achieves the best performance, especially in improving the smoothness and temporal consistency. 

\begin{figure}[!ht]
    \centering
     \includegraphics[
        width=\linewidth,
        trim=0 0.6cm 0 0,
        clip
    ]{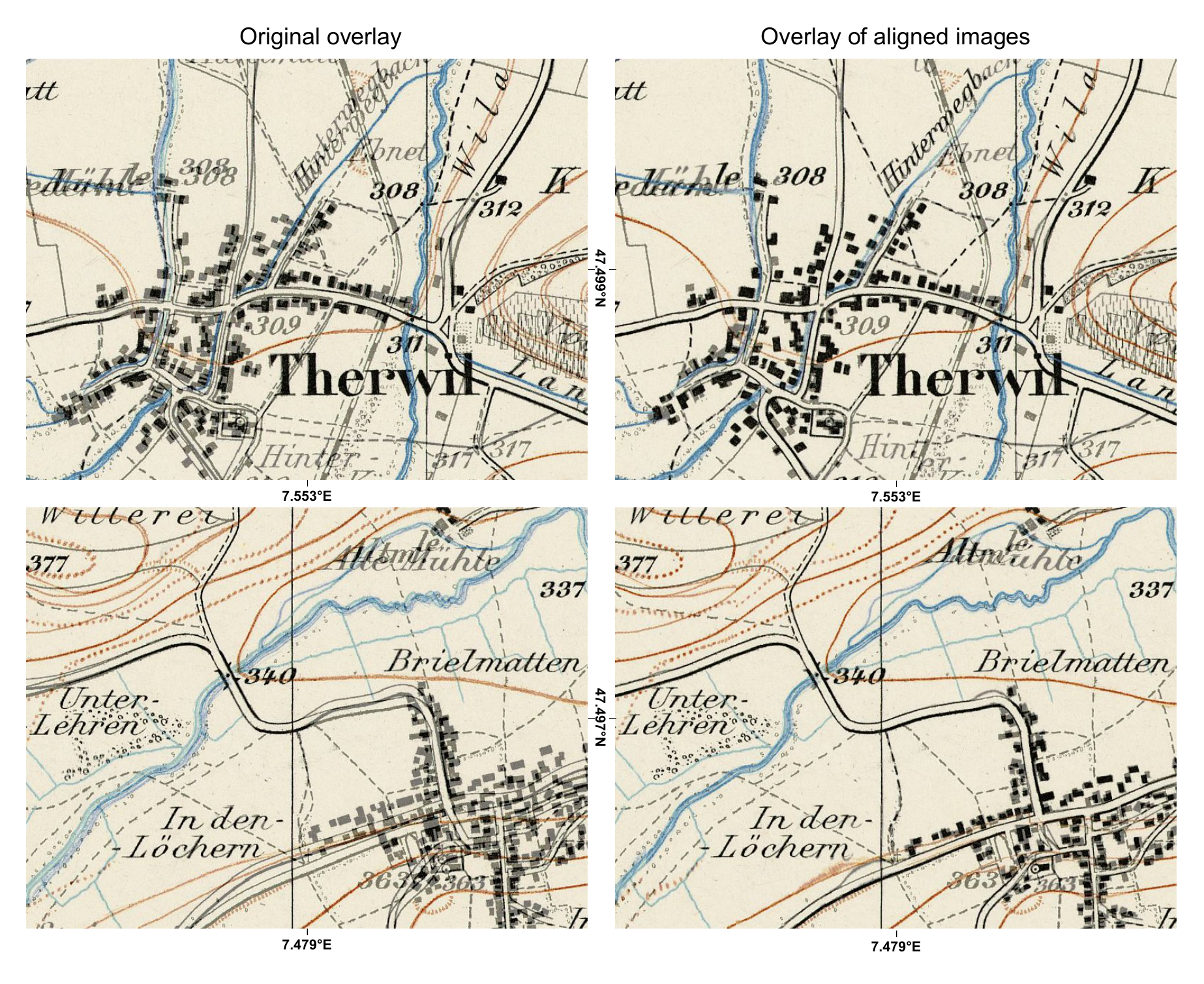}
    \caption{We re-trained our alignment model
    on Siegfried maps, a Swiss topographic map
    series at a scale of 1:25,000
    (\textcopyright{} swisstopo).
    Our model achieves plausible results in
    aligning various types of landscape features,
    highlighting its generalizability across
    different map types and scales.}

    \label{fig:siegfried_results}
\end{figure}

\begin{figure}[!ht]
    \centering
    \includegraphics[
        width=\linewidth,
        trim=0 0.6cm 0 0,
        clip
    ]{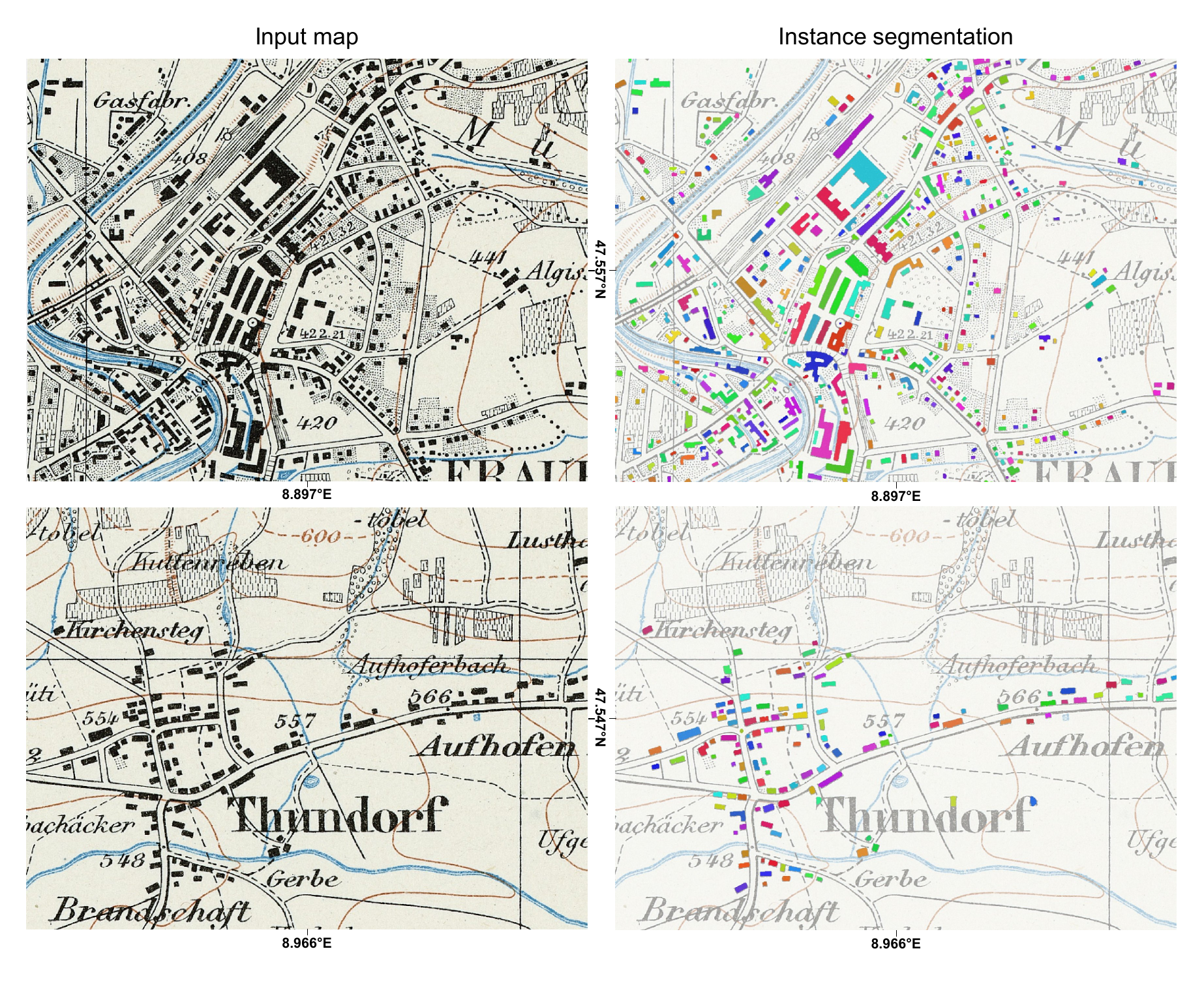}

    \caption{We re-trained our instance
    segmentation models on Siegfried maps
    (\textcopyright{} swisstopo).
    Different colors represent different
    instances. Our model achieves plausible
    results in extracting building instances.}

    \label{fig:siegfried_results_instance}
\end{figure}

We also re-trained the instance segmentation model on Siegfried maps. We crop maps into 256$\times$256 pixels. To reduce aliasing effects on small buildings caused by the fourfold downsampling and subsequent upsampling in the original Mask2Former architecture, we upsampled the map images by a factor of four during pre-processing. The model was trained on six map sheets, each of 7000$\times$4800 pixels. For temporal fusion, we additionally used the two neighbouring editions of each sheet, warped to the target edition using the predicted displacement field. For testing, we used four map sheets, each available in two editions, and processed them using the same warping strategy. Since the maps represent building footprints rather than block units, we adapted the definitions of small'' and large'' buildings to approximately 80 m$^2$ and 300 m$^2$, respectively, based on the building-area statistics of the \emph{Siegfried maps}. Both the Mask2Former baseline and our fusion strategy perform similarly well in instance segmentation, achieving considerably higher AP scores than in the Paris case due to the relatively lower complexity of building recognition. \Cref{fig:siegfried_results} and \Cref{fig:siegfried_results_instance} show the qualitative results of alignment and instance segmentation from our proposed models.

\begin{table}[h]
\caption{Model performance for displacement estimation on \emph{Siegfried maps}. Smoothness is evaluated by mV, and temporal consistency by L1 within triplets from consecutive years. Arrows indicate the direction of better performance}
\begin{tabular}{l c c c c c c}
\toprule
 & \textbf{Image} & \multicolumn{3}{c}{\textbf{Object}} & \textbf{Disp.} & \textbf{Temp. Cons.} \\
\cmidrule(lr){2-2} \cmidrule(lr){3-5} \cmidrule(lr){6-6} \cmidrule(lr){7-7}
\textbf{Method} 
& SSIM$\uparrow$ 
& CD(10\%)$\downarrow$ 
& CD(20\%)$\downarrow$ 
& CD(40\%)$\downarrow$ 
& mV$\downarrow$ 
& L1$\downarrow$ \\
\midrule
Unaligned   & 0.50 & 1.54 & 0.97 & 0.45 & --   & --   \\
GLU-Net      & 0.58 & 1.91 & 1.10 & 0.49 & 0.25 & 4.15 \\
WarpC        & 0.60 & 1.64 & 0.88 & 0.32 & 0.19 & 1.92\\

\textbf{Ours}
             & \textbf{0.61} & 1.65 & \textbf{0.88} & \textbf{0.31} & \textbf{0.17} & \textbf{1.79} \\
\bottomrule
\end{tabular}
\label{tab:additional_benchmark_alignment}
\end{table}

\begin{table}[h]
    \centering
    \caption{Average precision score for different objects categorized by their areas: small (area$\leq$80 m$^2$), medium (80 m$^2$<area<300 m$^2$), large (area>300 m$^2$). We compare using single maps (Mast2former as the baseline) and using our proposed multi-map fusion strategy.}
    \begin{tabular}{l c c c c }
    \toprule
    \textbf{Method} & AP\_{avg} & AP\_{s} & AP\_{m} & AP\_{l} \\ \midrule
    Mask2Former (single-map) &  0.87  & 0.27 & 0.92 & 0.87 \\ \midrule
    Ours (multi-map fusion)    & 0.86 & 0.30 & 0.87 & 0.85  \\ \bottomrule
    \end{tabular}
    \label{tab:instance_seg_sieg}
\end{table}

% ========== Figure 2: Segmentation ==========

% \begin{figure}[!ht]
%     \centering
%     \begin{tabular}{@{}cc@{}}
%         Input map &
%         Instance segmentation \\[1mm]

%         \includegraphics[
%             width=0.45\textwidth,
%             keepaspectratio
%         ]{siegfried1.jpg}
%         &
%         \includegraphics[
%             width=0.45\textwidth,
%             keepaspectratio
%         ]{siegfried1_instance.jpg}
%         \\[1mm]

%         \includegraphics[
%             width=0.45\textwidth,
%             keepaspectratio
%         ]{siegfried2.jpg}
%         &
%         \includegraphics[
%             width=0.45\textwidth,
%             keepaspectratio
%         ]{siegfried2_instance.jpg}
%     \end{tabular}

%     \caption{We re-trained our instance
%     segmentation models on Siegfried maps
%     (\textcopyright{} swisstopo).
%     Different colors represent different
%     instances. Our model achieves plausible
%     results in extracting building instances.}

%     \label{fig:siegfried_results_instance}
% \end{figure}

%\section*{Supplementary Note 7: Additional results on other maps}

\section*{Supplementary Note 7: Zurich change profiling}
\addcontentsline{toc}{section}{Supplementary Note 7: Zurich change profiling}

\begin{figure}[htbp]
    \centering
    \vspace{-0.3cm}
    % ================= Row 1 =================
    \begin{subfigure}{\textwidth}
        \centering
        \includegraphics[width=0.9\linewidth, trim=10 10 10 10, clip]{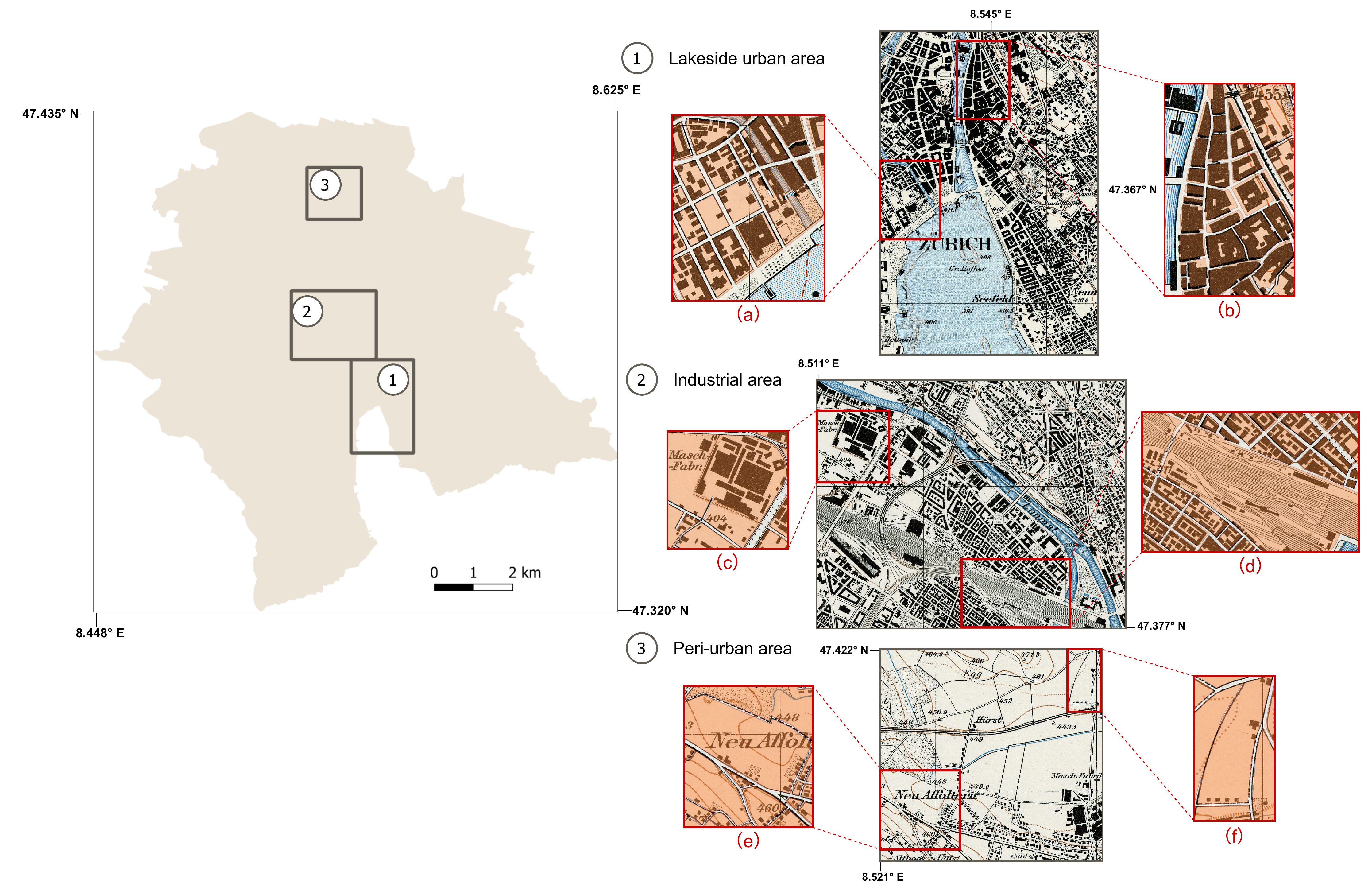}
        \vspace{-0.2cm}
        \caption{Overview of the Zurich case study.  Building blocks are represented as light orange polygons}
        \label{fig:zurich_overview}
    \end{subfigure}

    % ================= Row 2 =================
    \begin{subfigure}{0.8\textwidth}
        \centering
        \begin{minipage}{0.25\textwidth}
            \centering
            \includegraphics[width=\linewidth]{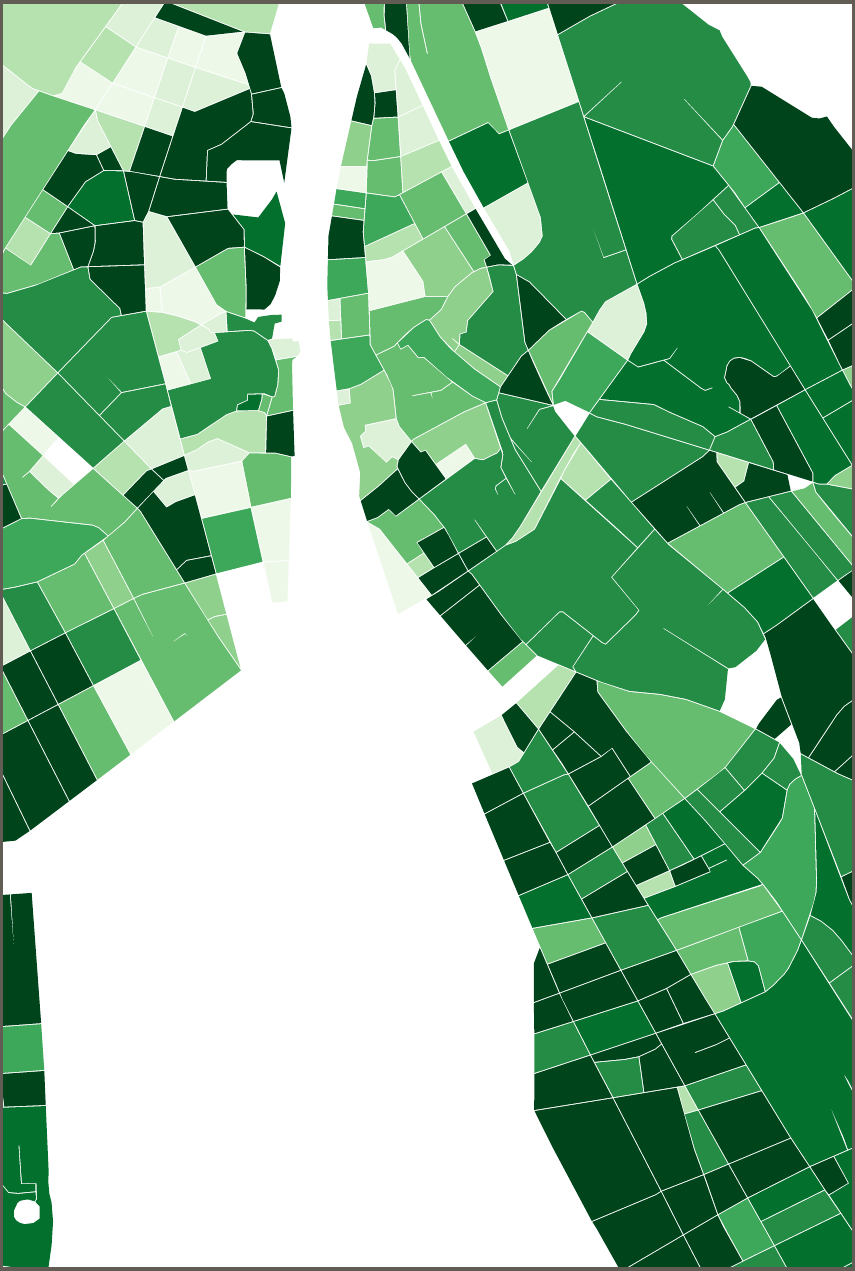}
        \end{minipage}
        \hspace{0.01\textwidth}
        \begin{minipage}{0.25\textwidth}
            \centering
            \includegraphics[width=\linewidth]{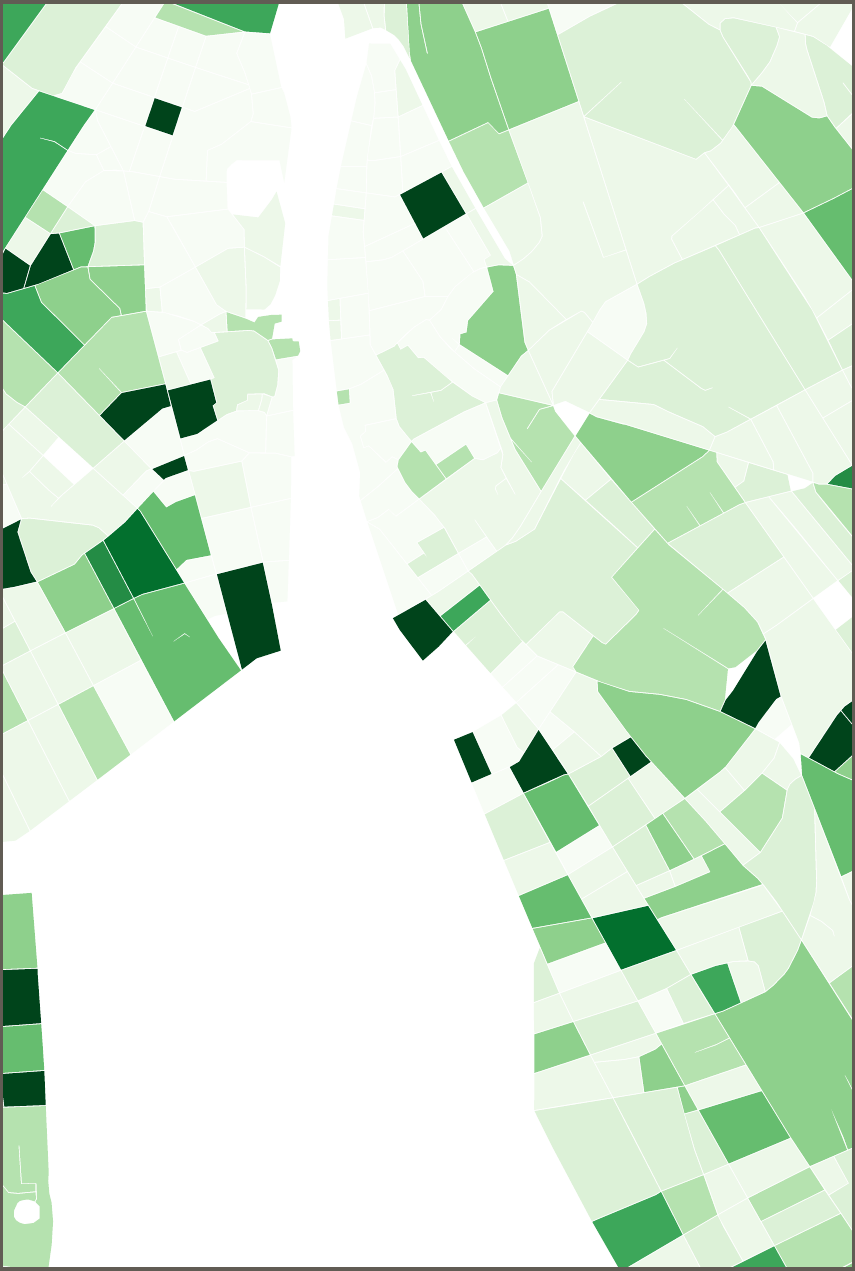}
        \end{minipage}
        \hspace{0.01\textwidth}
        \begin{minipage}{0.25\textwidth}
            \centering
            \includegraphics[width=\linewidth]{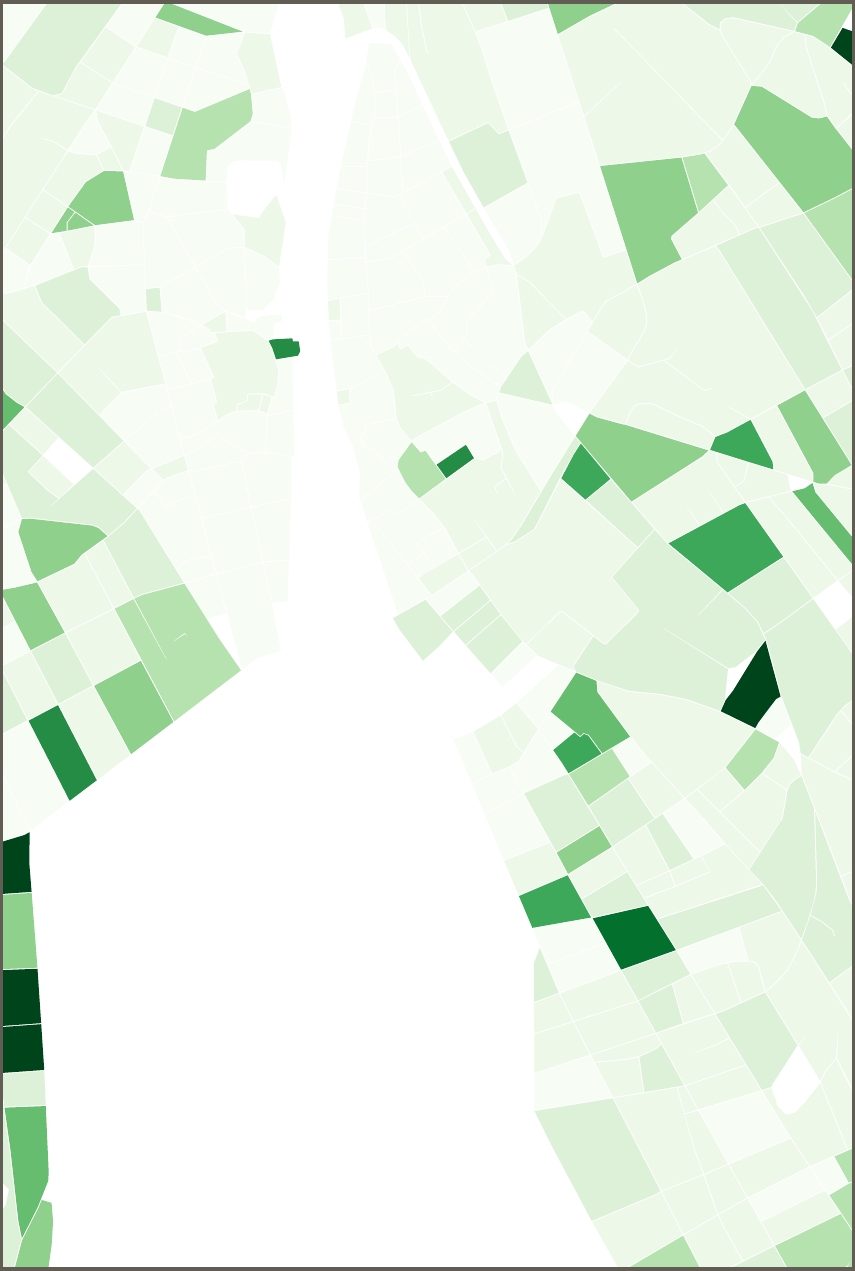}
        \end{minipage}
        \hspace{0.01\textwidth}
        % ===== Right: vertical colorbar =====
        \begin{minipage}{0.04\linewidth}
            \centering
            \includegraphics[height=0.8\linewidth, angle=90]{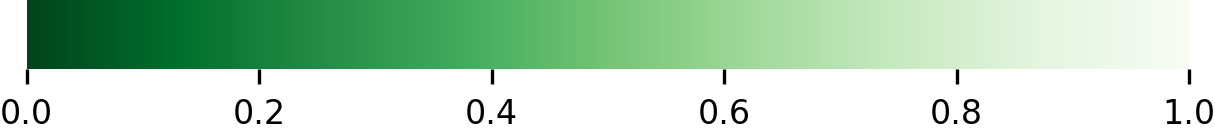}
            \caption*{\rotatebox{90}{IoU}}
        \end{minipage}
        \vspace{-0.5mm}
        \caption{Urban changes in \textcircled{1} of 1893-1913, 1913-1932, 1932-1940.}
        \label{fig:zurich_lakeareas}
    \end{subfigure}

    % ================= Row 3 =================
    \begin{subfigure}{0.9\textwidth}
        \centering
        \begin{minipage}{0.3\textwidth}
            \centering
            \includegraphics[width=\linewidth]{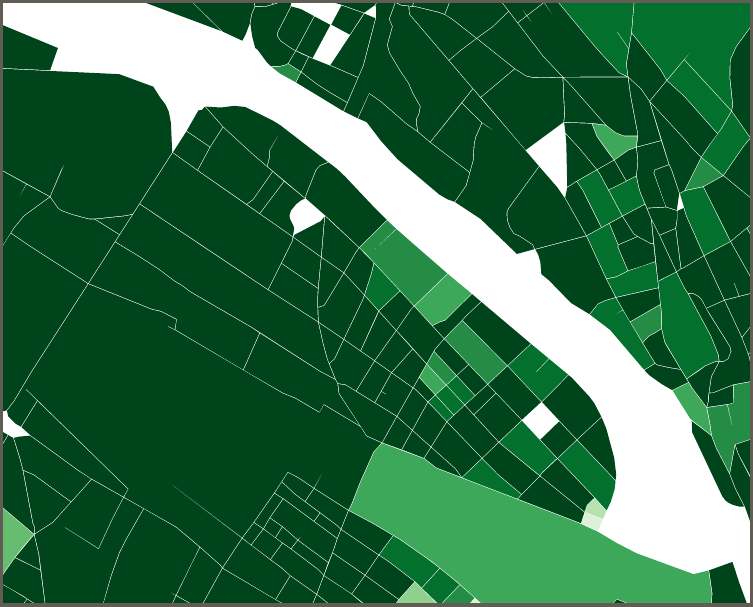}
        \end{minipage}
        \hspace{0.01\textwidth}
        \begin{minipage}{0.3\textwidth}
            \centering
            \includegraphics[width=\linewidth]{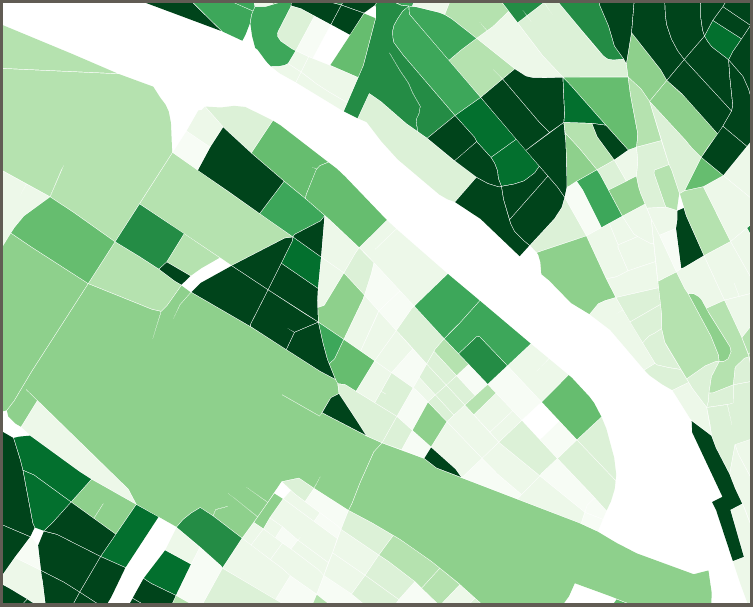}
        \end{minipage}
        \hspace{0.01\textwidth}
        \begin{minipage}{0.3\textwidth}
            \centering
            \includegraphics[width=\linewidth]{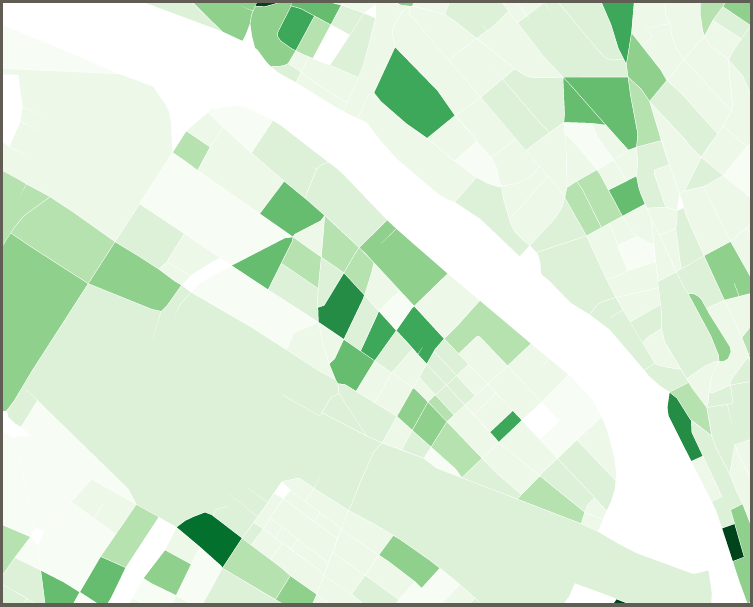}
        \end{minipage}
        \vspace{-0.5mm}
        \caption{Urban changes in \textcircled{2} of 1893-1913, 1913-1932, 1932-1940.}
        \label{fig:zurich_railway}
    \end{subfigure}

    % ================= Row 4 =================
    \begin{subfigure}{0.8\textwidth}
        \centering
        \begin{minipage}{0.28\textwidth}
            \centering
            \includegraphics[width=\linewidth]{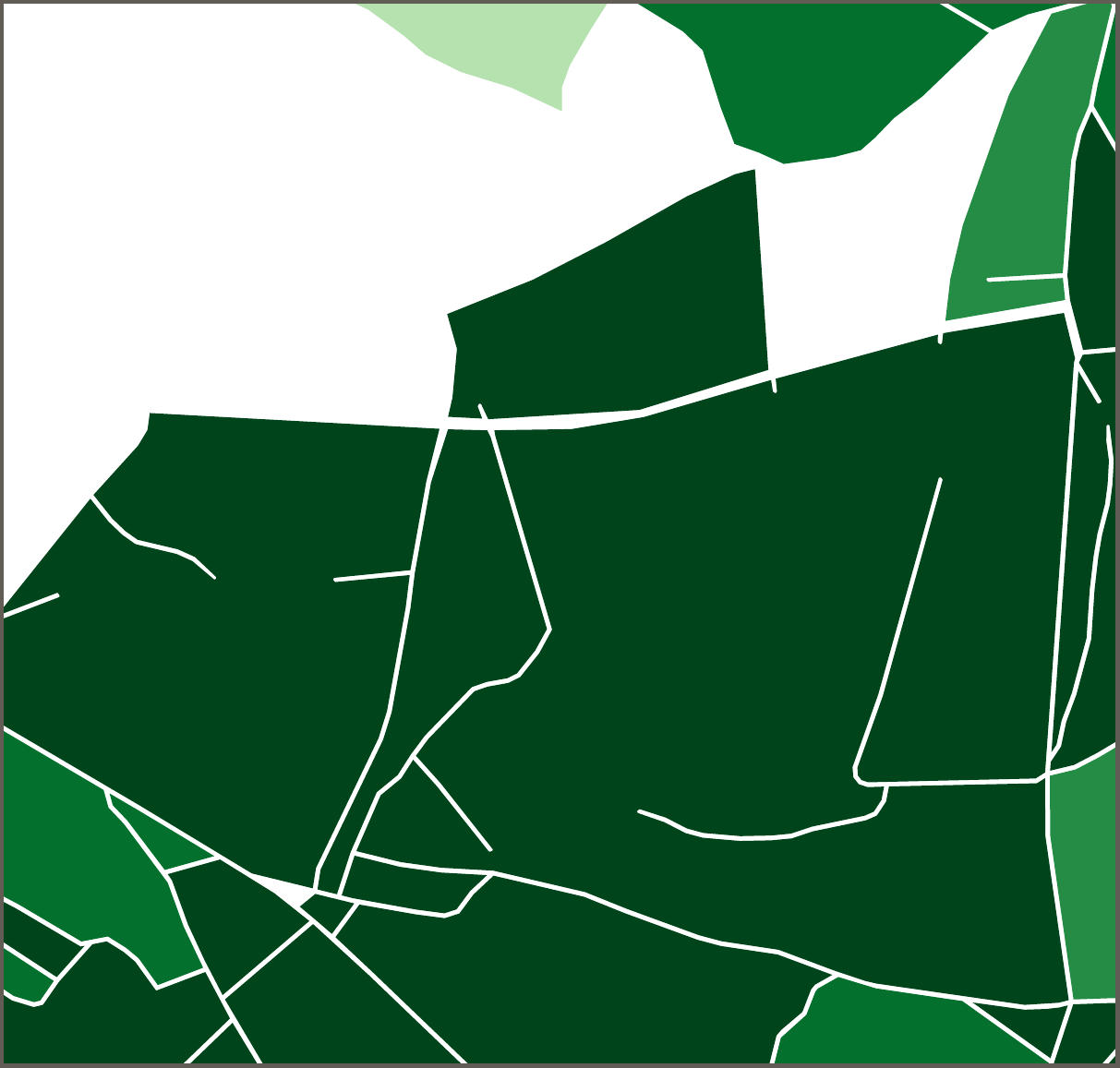}
        \end{minipage}
        \hspace{0.01\textwidth}
        \begin{minipage}{0.28\textwidth}
            \centering
            \includegraphics[width=\linewidth]{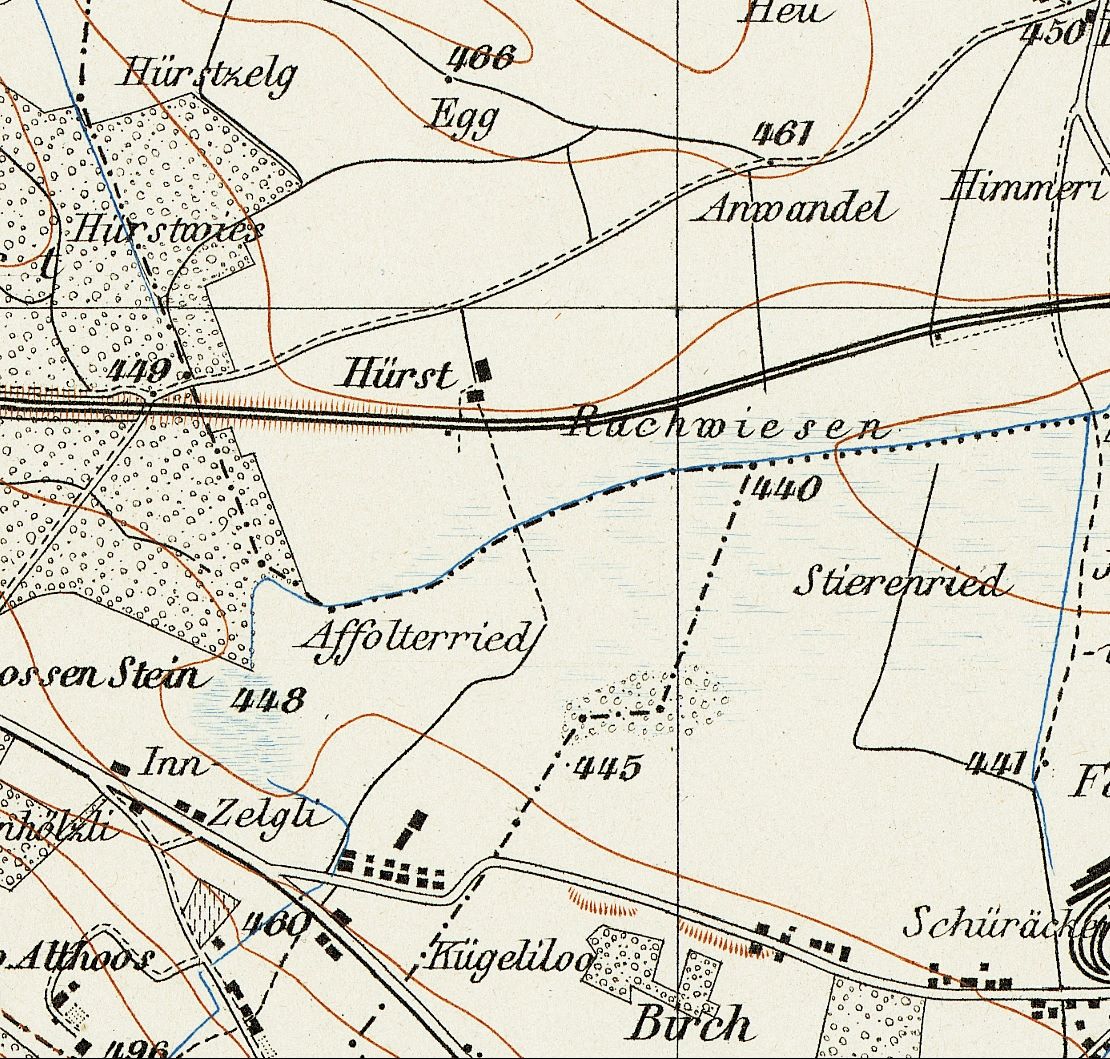}
        \end{minipage}
        \hspace{0.01\textwidth}
        \begin{minipage}{0.28\textwidth}
            \centering
            \includegraphics[width=\linewidth]{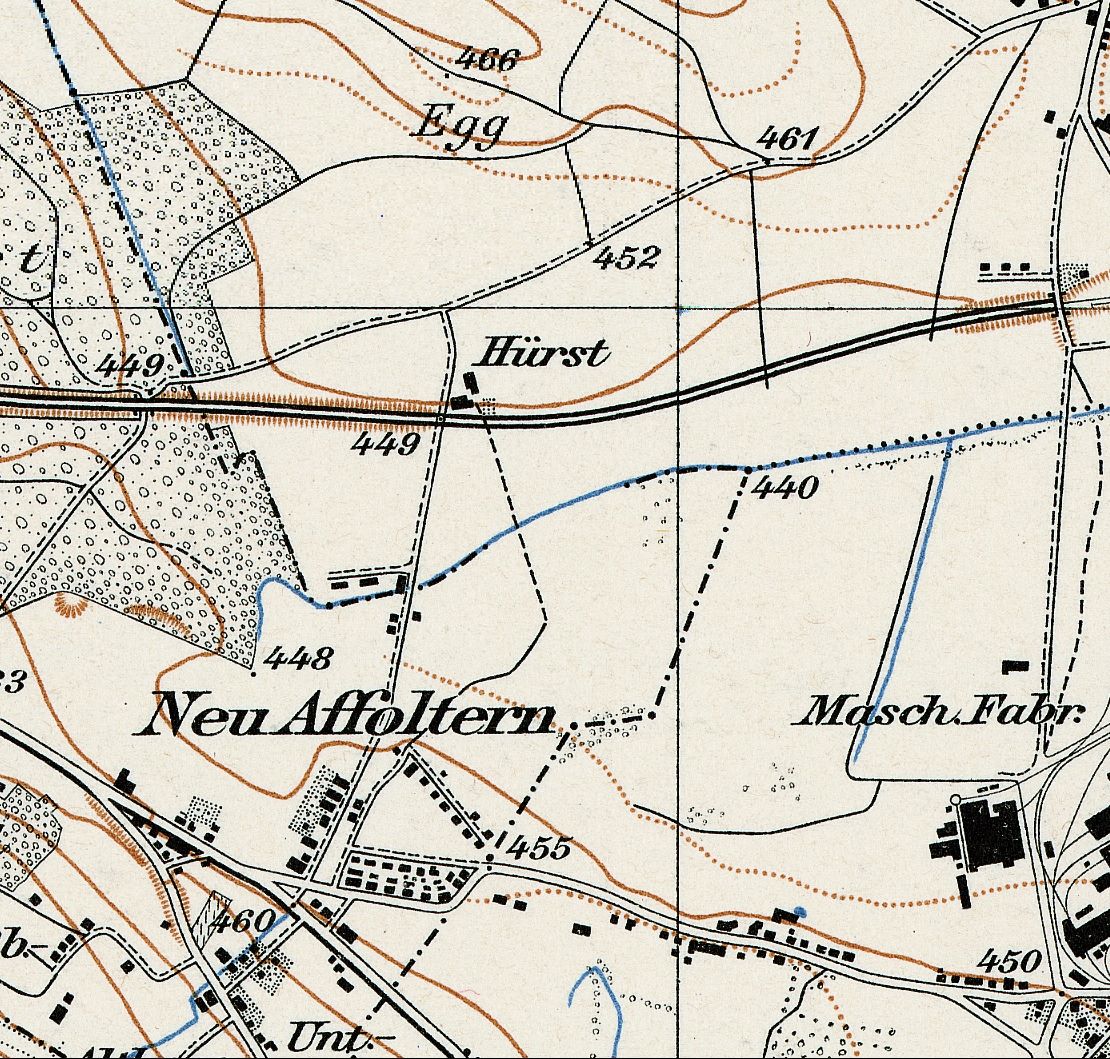}
        \end{minipage}
        \vspace{-0.5mm}
        \caption{Urban development in \textcircled{3} between 1896 and 1913 and original maps.}
        \label{fig:zurich_neuaffoltern}
    \end{subfigure}
    \vspace{-0.25cm}
    \caption{Zurich change profiling on Siegfried maps (\textcopyright{} swisstopo), 1893–1940. (\subref{fig:zurich_overview}) Overview of the Zurich case-study areas, including \textcircled{1} lakeside urban areas characterized by dense residential urban blocks; \textcircled{2} industrial areas (``Industriequartier'') and the main railway station dominated by large industrial and railway infrastructures; and \textcircled{3} peri-urban areas with sparse road networks and scattered buildings. (\subref{fig:zurich_lakeareas}) Block-level change profiles for dense lakeside urban areas. (\subref{fig:zurich_railway}) Block-level change profiles for the industrial and railway-station area. (\subref{fig:zurich_neuaffoltern}) Peri-urban example showing the limitation of road-bounded block definitions in sparse settlement contexts. Darker colours indicate stronger detected change.}
    \label{fig:change_zurich}

\end{figure}

We test the complete change profiling pipeline on the urban development of Zurich between 1893 and 1940 using Siegfried maps at the scale of 1:25,000. Three representative case study areas are selected, as shown in~\Cref{fig:change_zurich}: \protect\textcircled{1} lakeside urban areas characterized by dense residential urban blocks; \protect\textcircled{2} industrial areas (``Industriequartier'') and the main railway station dominated by large industrial and railway infrastructures; and \textcircled{3} peri-urban areas with sparse road networks and scattered buildings.

For block formation, we use the same definition as~\cite{chazalon2021icdar} for Paris maps.
For Siegfried maps, blocks are represented differently across urban contexts.
In \textcircled{1}(a) and \textcircled{2}(c)(d), the blocks are explicitly delineated by roads, containing smaller buildings separated by open spaces. In \textcircled{1}(b), especially in the old-town fabric, the compact built-up masses themselves correspond to block units (shown in large black polygons), separated by narrow street spaces.
Despite these differences, the mapped units mainly capture block-level built form rather than fine-grained building details. Therefore, the Siegfried maps are less suitable for micro-level structural monitoring, but remain meaningful for capturing broader patterns of urban form and morphological change.

As shown in~\cref{fig:zurich_lakeareas}, the lakeside urban areas underwent substantial transformations, particularly on the eastern lakeshore between 1893 and 1932, while remaining comparatively stable afterward. 
%This pattern reflects the intense modernization, residential expansion, and parcel reorganization associated with Zurich’s metropolitan growth during the late \nth{19} and early \nth{20} centuries. 
This pattern is consistent with Zurich’s rapid metropolitan growth after the 1893 municipal incorporation, when new planning instruments such as the \emph{Quartierplan} enabled the reorganization of land ownership, land use, public space, and urban form at the neighbourhood scale~\cite{kattenbeck2025urban}.
Afterward, the detected changes became increasingly incremental, suggesting a relative stabilization of the urban structure.
In~\cref{fig:zurich_railway}, the industrial and railway station districts also experienced major transformations during 1893--1913, reflecting Zurich’s broader industrial-era development that includes railway expansion, industrialization, and the construction of new factories and logistics infrastructures~\cite{herger2023railway}. The pace of transformation slowed during 1913--1932, while only sporadic modifications occurred between 1932 and 1940.
These observations highlight the significant urbanization processes that shaped Zurich during the late \protect \nth{19} and early \protect \nth{20} centuries, characterized by intensive residential, industrial, and infrastructural expansion.

The results further demonstrate that the proposed change profiling framework performs reasonably well even on topographic maps of a smaller scale compared to city maps in urban environments such as \textcircled{1} and \textcircled{2}, where infrastructure-bounded (primarily road-bounded) urban blocks constitute meaningful morphological entities. However, the same framework is not directly transferable to peri-urban and suburban landscapes characterized by sparse infrastructures and discontinuous settlement patterns. As illustrated in \textcircled{3}, applying the same Paris-derived definition of urban blocks leads to excessively large polygons because of the low density of roads. In such contexts, the resulting units are topologically valid but morphologically weak, as they no longer correspond to coherent urban blocks. Additional parcel structures and infrastructural divisions can provide more meaningful units for organizing and interpreting change; alternatively, changes can be analyzed directly at the building level where individual buildings are clearly represented.
Moreover, since only a limited number of buildings are contained within each block, even minor building modifications may produce disproportionately strong change signals, as depicted in~\Cref{fig:zurich_neuaffoltern}. In this case, the detected changes become less informative and less reliable for interpreting actual urban transformation processes at the block level, but may still be meaningful when interpreted at the localized building-level.

\section*{Supplementary Note 8: Cross-scale and cross-series alignment}
\addcontentsline{toc}{section}{Supplementary Note 8: Cross-scale and cross-series alignment}
To assess the applicability of the framework beyond a single homogeneous map series, we further examine comparisons across scales and across cartographic series. Such situations arise naturally when additional map sources are introduced to extend temporal coverage beyond the span of a single internally consistent product. In these settings, the alignment module may still remain useful, but only if the compared maps preserve sufficient spatial correspondence and structural similarity for dense matching.

\begin{figure}[H]
    \centering
    \vspace{-0.2cm}
    \includegraphics[width=0.9\linewidth, trim=0 20 0 0, clip]{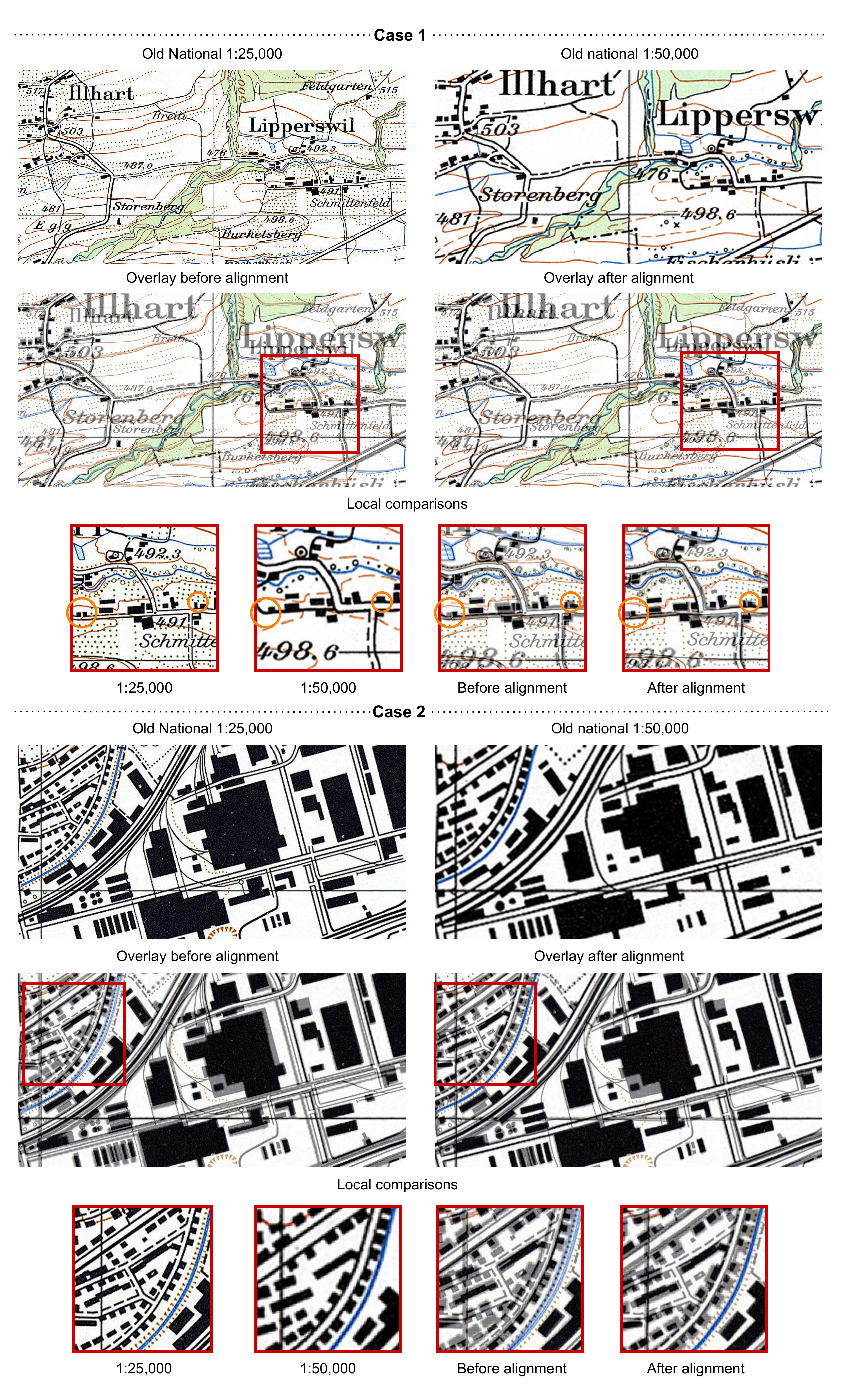}
    \caption{Cross-scale performance. We test the alignment performance across Old National maps of 1:25,000 and 1:50,000 of the same period (\textcopyright{} swisstopo). While the alignment module is still effective, change profiling is no longer reliable due to generalization difference cross scales.}
    \label{fig:cross-scale}
\end{figure}

In~\Cref{fig:cross-scale}, we present alignment results for two maps from the \emph{Old National Maps} series at different scales, 1:25,000 and 1:50,000, from the same time period, provided by swisstopo~\cite{national_map}.  This experiment isolates the effect of scale without introducing real temporal change. Although the alignment module can still partially normalize the positions of major features, instance-level comparison becomes less meaningful because the two scales encode different levels of detail through scale-dependent cartographic generalization. We showcase two examples: Case~1 for peri-urban areas and Case~2 for dense urban areas. As highlighted by the orange circles in the local comparison for Case~1, two individual buildings at 1:25,000 may be merged into a single larger building representation at 1:50,000. In Case~2, the building blocks shown in the zoomed-in panels are similarly simplified at the coarser scale. This experiment shows that geometric alignment may remain feasible across scales, whereas direct downstream change quantification is no longer directly interpretable.

\begin{figure}[H]
    \centering
    \includegraphics[width=0.9\linewidth, trim=10 10 10 10, clip]{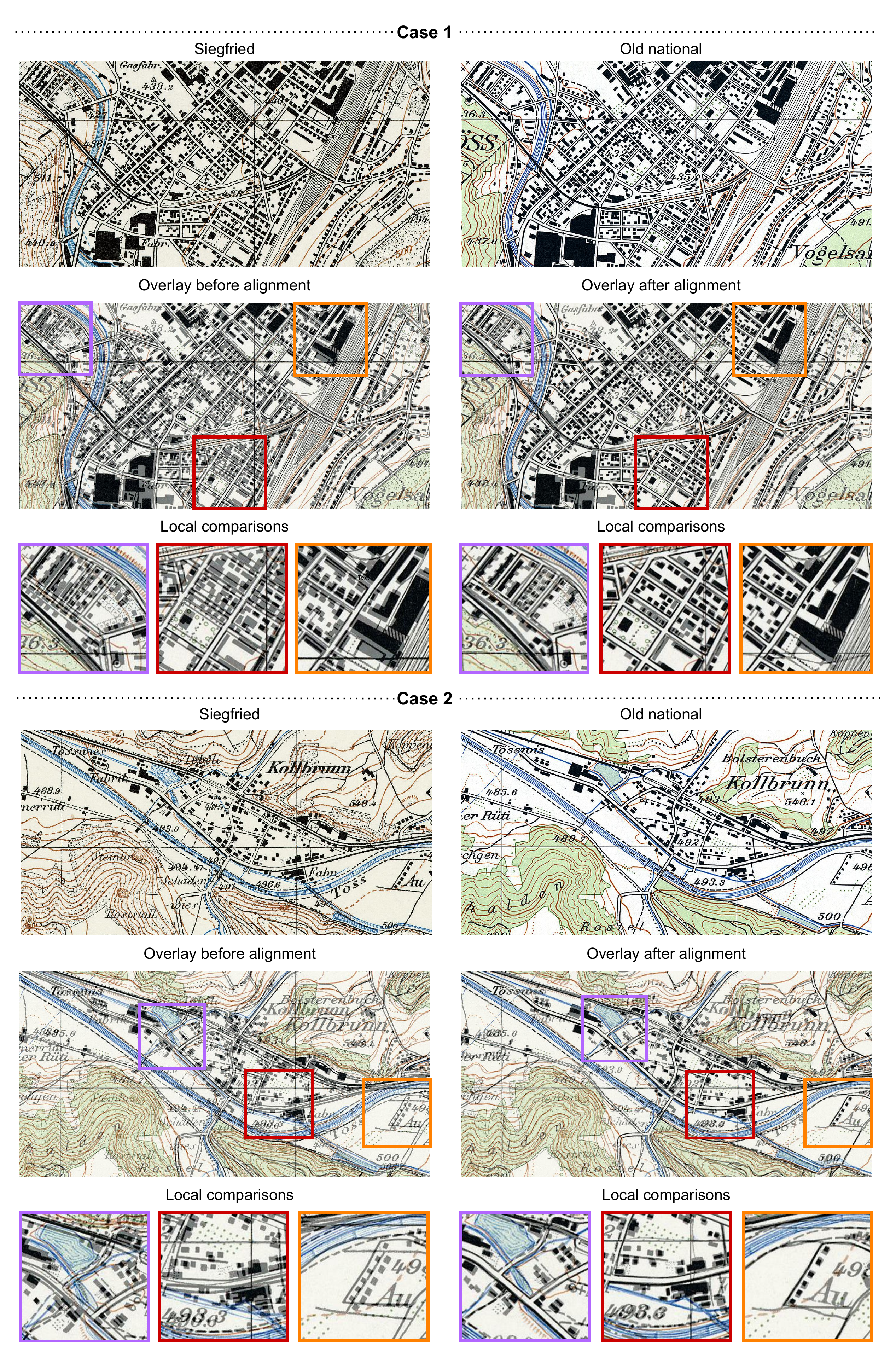}
    \caption{Cross-series performance. We test the alignment performance across Siegfried maps (1880s--1940s) and Old National maps (1950s--1990s), both at the scale of 1:25,000 (\textcopyright{} swisstopo). The alignment module remains effective and the generalization level remains consistent, altogether enabling the urban analysis at a larger temporal span.}
    \label{fig:cross-style}
\end{figure}

In~\Cref{fig:cross-style}, we apply the model to two different map series at the same nominal scale, \emph{Siegfried} and \emph{Old National Maps}, both at 1:25,000. We present two examples: one from a dense urban area (Case~1) and the other from a peri-urban area (Case~2). These series exhibit noticeable differences in symbolization and contour rendering, while still preserving sufficient visual and spatial correspondence, particularly for roads, rivers, railways, and settlement patterns. Under this condition, the alignment module remains effective, especially over built-up areas. Since \emph{Siegfried} covers the 1880s--1940s and \emph{Old National Maps} the 1950s--1990s, this setting also suggests that alignment-based comparison may help connect related topographic series across longer historical periods, potentially extending analysis beyond the temporal span of a single map product.

Taken together, these experiments indicate that the alignment module can generalize beyond a single homogeneous map series when the compared maps still retain sufficient common geometric structure for dense matching. By contrast, the downstream change-analysis stage is more restrictive, since its interpretability depends not only on geometric correspondence but also on preserving comparable semantic granularity across editions. In practice, this means that heterogeneous sources may still support alignment-based comparison, while fine-grained change quantification remains most reliable within map series that are internally consistent in both scale and representation.

\section*{Supplementary Note 9: Applications of displacement estimation for map distortion studies}
\addcontentsline{toc}{section}{Supplementary Note 9: Applications of displacement estimation for map distortion studies}

\begin{figure}[H]
    \centering
    \includegraphics[width=0.85\linewidth]{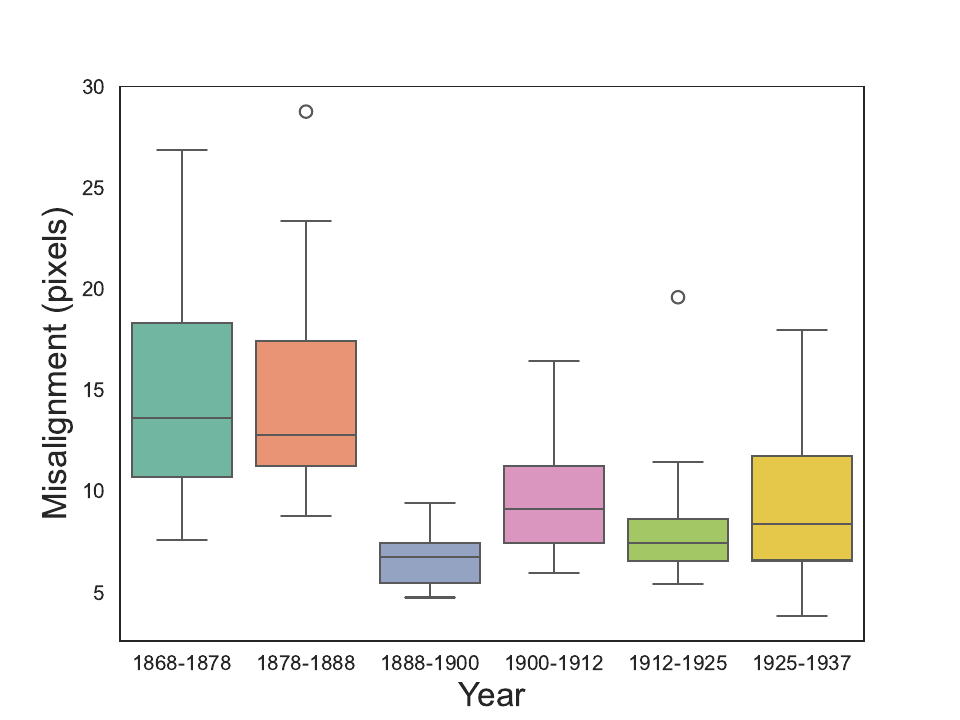}
    \caption{Misalignment (in pixels) between maps over time. For each region, we compute the median of estimated per-pixel displacement, and the boxplot summarizes their distribution across all regions. This illustrates the potential of displacement estimation for analyzing map distortions and the evolution of cartographic techniques. }
    \label{fig:misalign_over_time}
\end{figure}

\section*{Supplementary Note 10: Additional figures on label propagation}
\addcontentsline{toc}{section}{Supplementary Note 10: Additional figures on label propagation}
\begin{figure}[H]
    \centering
    \includegraphics[width=0.9\linewidth]{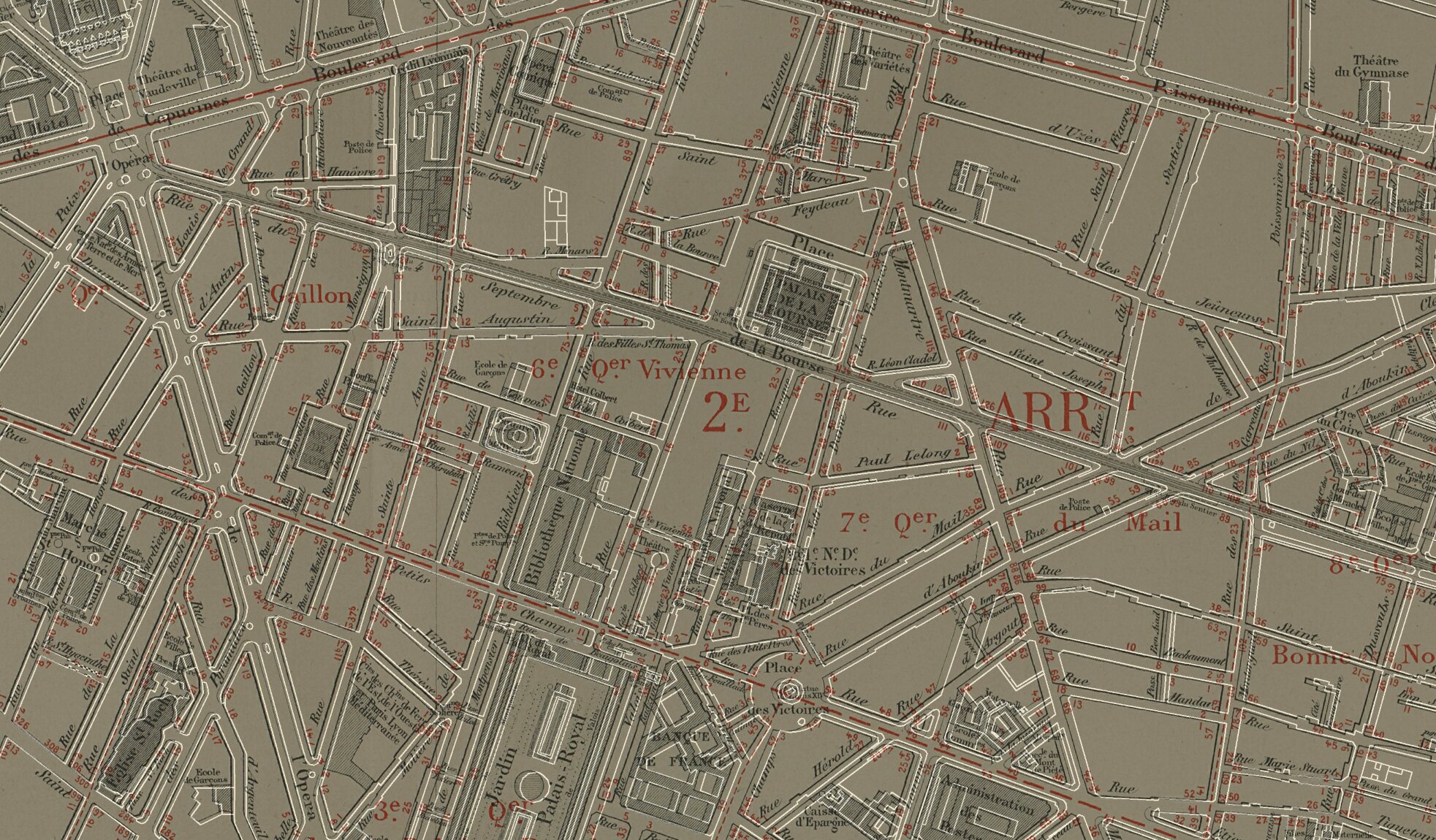}\\
    \includegraphics[width=0.9\linewidth]{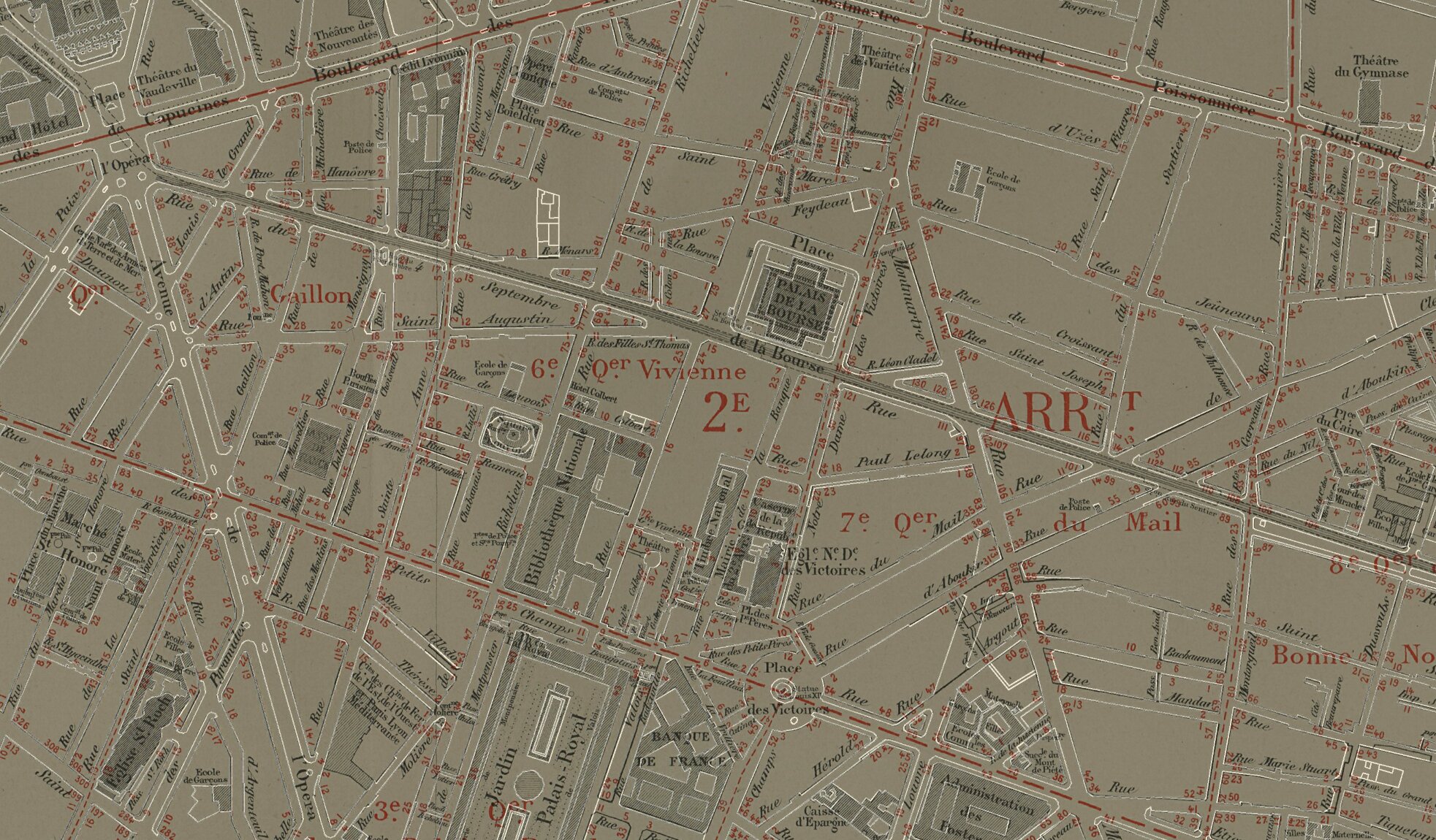}
    \caption{Example of propagated labels from 1925 to 1905, shown in edge maps~\citep{villeparis1925atlas}. Top: the original edge map from 1925, which exhibits clear misalignment with the map from 1905. Bottom: the edge map propagated using the estimated displacement field from 1925 to 1905. The propagated labels align closely with the target map, illustrating the ability of the proposed method to transfer annotations accurately across decades despite substantial urban change and cartographic variation.} 
    \label{fig:label_propagation}
\end{figure}

% \begin{figure}[H]
%     \centering

%     \mapwithcoords{overlay_ori_compress.png}

%     \par\medskip

%     \mapwithcoords{overlay_warp_1905_compress.png}

%     \caption{
%         Example of propagated labels from 1925 to 1905,
%         shown in edge maps~\citep{villeparis1925atlas}.
%         Top: the original edge map from 1925.
%         Bottom: the edge map propagated using the
%         estimated displacement field from 1925 to 1905.
%     }

%     \label{fig:label_propagation}
% \end{figure}

\section*{Supplementary Note 11: Sensitivity to temporal gaps}
\addcontentsline{toc}{section}{Supplementary Note 11: Sensitivity to temporal gaps}

\begin{figure}[ht]
    \centering
    \includegraphics[width=\linewidth]{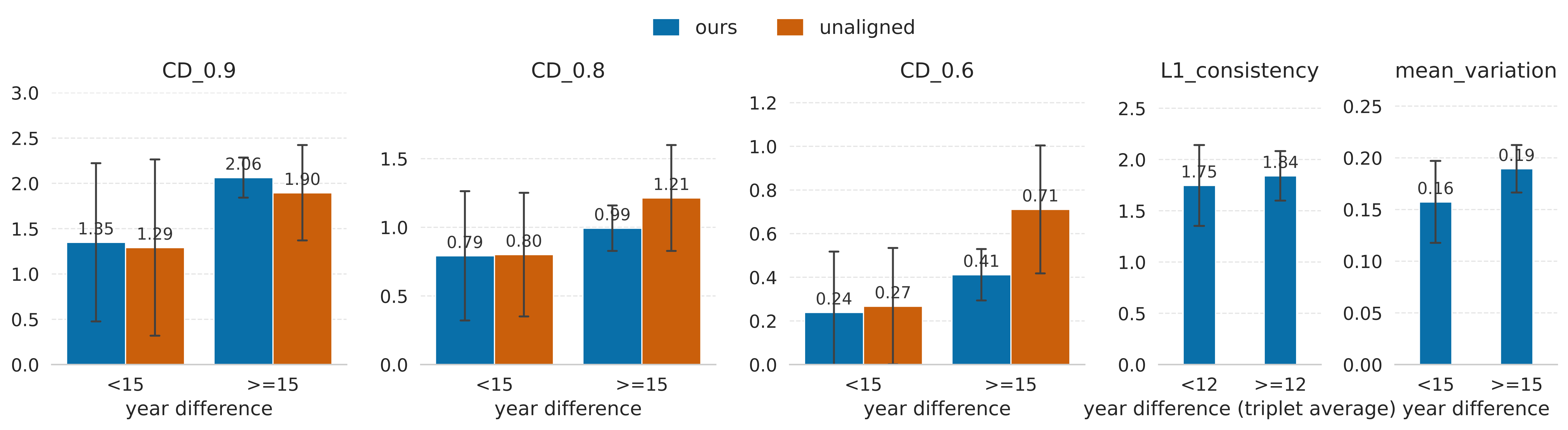}
    \caption{Temporal sensitivity analysis. We evaluate alignment performance on 12 map-edition pairs with different temporal gaps, grouped into smaller-gap (<15 years) and larger-gap ($\geq$15 years) subsets with 7 and 5 pairs, respectively. }
    \label{fig:temporal_sensitivity}
\end{figure}

We additionally assess temporal sensitivity on \emph{Siegfried map} pairs with varying inter-edition gaps. Unlike the \emph{Atlas Municipal} series, which presents roughly uniform temporal intervals between editions, consecutive Siegfried editions can differ by less than 10 years or by more than 20 years. We therefore group the 12 test pairs into smaller-gap (<15 years) and larger-gap ($\geq$15 years) subsets with 7 and 5 pairs, respectively. As shown in~\Cref{fig:temporal_sensitivity}, alignment quality degrades moderately for larger-gap pairs, as indicated by slightly increased L1 consistency discrepancy and mean variation. The CD of the unaligned maps is also higher for larger-gap pairs, likely reflecting a combination of stronger genuine map changes and larger geometric discrepancies. Even so, the proposed alignment method remains effective in both settings, and is particularly beneficial for larger-gap pairs, where it reduces both the mean CD and its standard deviation relative to the unaligned baseline.

These results suggest that temporally closer editions are generally preferable for fine-grained alignment and downstream change quantification, although we do not observe a sharp year-based failure threshold within the range examined here. 
We only observe a few rare failure cases, in which the cartographic representation differs substantially even between nearby editions, for example in the depiction of contour lines. Such cases likely arise because some topographic features were manually interpolated and redrawn according to edition-specific drafting conventions. Overall, these exceptions indicate that the practical boundary of the method is governed not by temporal gap alone, but by the extent to which the compared editions still preserve comparable spatial structure and semantic content.

\end{appendices}

% \bibliographystyle{plain}
% \bibliography{references}

\section*{Acknowledgments}

The research was initiated during S.W.’s research exchange at LASTIG, IGN, and was subsequently conducted at the Institute of Cartography and Geoinformation at ETH Zurich in close collaboration with LASTIG, IGN on the SoDUCo project ``Social Dynamics in Urban Context'' [ANR-18-CE38-0013].

\section*{Funding statement}
The project was supported by the Swiss National Science Foundation (SNSF), as the part of the project ``HistoRiCH: Historical river change - Planning for the future by exploring the mapped past'' [No. 188692]. The open access funding was provided by ETH Zurich.

\section*{Author contributions}
S.W., Y.C., and J.P. conceptualized the study. S.W. processed the data, developed the methodology, implemented the workflow, conducted the experiments and analyses, visualized the results, and wrote the manuscript. Y.C. contributed to data processing and supported the implementation. M.G. validated the IoU-based change maps and provided historical insights. K.S. and C.M. provided technical guidance and feedback on the manuscript. J.P. contributed to the analysis, provided historical insights, and coordinated the project. L.H. provided computational resources and feedback on the project. Both J.P. and L.H. supervised the project.
All authors reviewed and edited the manuscript.

\section*{Competing interests}
The authors declare no competing interests.

\end{document}